\documentclass{article}
\usepackage[utf8]{inputenc}
\usepackage[small]{dgruyter}

\usepackage{draftfigure}

\usepackage{amsmath,amssymb,libertine,multicol,subcaption,lgreek,soul,csquotes}
\usepackage[many]{tcolorbox}
\usepackage{tikz,pgfplots,tikz-qtree}
\pgfplotsset{compat=1.16}
\usepackage[bottom]{footmisc}
\tcbset{colframe=gray}
\usepackage[style=authoryear-comp,dashed=false,sortcites=false,useprefix=true,hyperref=true,uniquename=false,maxcitenames=2,maxbibnames=10,isbn=false]{biblatex}
\DeclareCiteCommand{\citeyear}
    {\usebibmacro{prenote}}
    {\bibhyperref{\printfield{year}}\bibhyperref{\printfield{extrayear}}}
    {\multicitedelim}
    {}

\newcommand{\citeposs}[1]{\citeauthor{#1}'s (\citeyear{#1})}

\begin{document}
\begin{tcolorbox}[colback=yellow!70]
  \texttt{** NOTE: This is a pre-print of a version of this paper that has been accepted (Nov. 2021) for publication in \textit{Corpus Linguistics \& Linguistic Theory}. The official version will be linked in due time. ** }
\end{tcolorbox}
\author*[1]{Martijn van der Klis}
\author[2]{Jos Tellings}
\runningauthor{M. van der Klis and J. Tellings}
\affil[1]{UiL OTS, Utrecht University, The Netherlands}
\affil[2]{UiL OTS, Utrecht University, The Netherlands}
\title{Generating semantic maps through multidimensional scaling: linguistic applications and theory}
\runningtitle{Generating semantic maps through multidimensional scaling}

\abstract{
This paper reports on the state-of-the-art in application of multidimensional scaling (MDS) techniques to create semantic maps in linguistic research. MDS refers to a statistical technique that represents objects (lexical items, linguistic contexts, languages, etc.) as points in a space so that close similarity between the objects corresponds to close distances between the corresponding points in the representation. We focus on the use of MDS in combination with parallel corpus data as used in research on cross-linguistic variation.

We first introduce the mathematical foundations of MDS and then give an exhaustive overview of past research that employs MDS techniques in combination with parallel corpus data. We propose a set of terminology to succinctly describe the key parameters of a particular MDS application. 

We then show that this computational methodology is theory-neutral, i.e.\ it can be employed to answer research questions in a variety of linguistic theoretical frameworks. Finally, we show how this leads to two lines of future developments for MDS research in linguistics.
}

\keywords{parallel corpora, multidimensional scaling, semantic maps, cross-linguistic variation}

\articletype{Article}
\journalname{ArxiV - non-official}
\journalyear{2021}

\maketitle

\section{Introduction} \label{sec:intro}

Multidimensional scaling (henceforth MDS) is a statistical technique that represents objects (lexical items, linguistic contexts, languages, etc.) in a dataset as points in a multidimensional space so that close similarity between objects in the dataset corresponds to close distances between the corresponding points in the representation. Typically, MDS reduces a dataset that has variation in a large number of dimensions, to a representation in only two or three dimensions. MDS can therefore be seen as a dimensionality reduction technique, which facilitates the graphical representation of a highly complex dataset as a 2D or 3D scatter plot. We will call such a visualization obtained through MDS an \textit{MDS map}.\footnote{Author names are in alphabetical order to reflect a shared responsibility for the content of the paper. Van der Klis wrote sections \ref{sec:intro}, \ref{sec:dimension}, \ref{sec:cluster}, and \ref{sec:alternatives}. Tellings wrote sections \ref{sec:math}, \ref{sec:types}, \ref{sec:theory}, and \ref{sec:composition}. The revising and structuring of the paper as a whole is a collective effort by both authors.}

MDS as a statistical and visualization tool has been used in various fields of science (see e.g.\ \cite{Ding2018}). Recently, researchers in linguistic typology have started using MDS as a method to chart cross-linguistic variation in semantic maps (see \cite{Cysouw2001,Levinson2003} and especially \cite{Croft2008} for early work). Semantic maps visually represent interrelationships between meanings expressed in languages. In \citeposs{Anderson1982} typological work on the \textsc{perfect} construction, the idea of a semantic map was introduced as a method of visualizing cross-linguistic variation. In \citeposs{Haspelmath1997} work on indefinites, semantic maps were formalized as graphs: nodes display functions (or meanings) in the linguistic domain under investigation, while edges convey that at least one language has a single form to express the functions of the nodes the edge connects. 

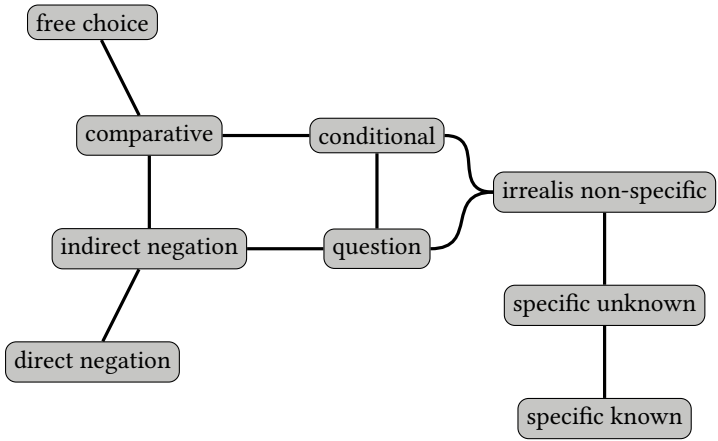
\begin{figure}[!ht]
\centering
\begin{tikzpicture}[scale=.75]
    \tikzstyle{mynode}=[fill={rgb,255: red,190; green,190; blue,190}, draw=black, shape=rectangle, rounded corners]

		\node [style=mynode] (1) at (-2, -7.25) {specific known};
		\node [style=mynode] (4) at (-2, -5.25) {specific unknown};
		\node [style=mynode] (6) at (-2, -3.25) {irrealis non-specific};
		\node [style=mynode] (9) at (-6, -4.25) {question};
		\node [style=mynode] (11) at (-6, -2.25) {conditional};
		\node [style=mynode] (12) at (-10, -4.25) {indirect negation};
		\node [style=mynode] (13) at (-10, -2.25) {comparative};
		\node [style=mynode] (14) at (-11, -6.25) {direct negation};
		\node [style=mynode] (16) at (-11, -.25) {free choice};

		\draw [very thick] (1) to (4);
		\draw [very thick] (4) to (6);
                \draw [very thick, in=360, out=-180, looseness=1.50] (6) to (9);
		\draw [very thick]  (9) to (12);
		\draw [very thick]  (12) to (14);
                \draw [very thick,in=360, out=-180, looseness=1.50] (6) to (11);
		\draw [very thick]  (11) to (9);
		\draw [very thick]  (11) to (13);
		\draw [very thick]  (13) to (12);
		\draw [very thick]  (13) to (16);

\end{tikzpicture}
\caption{Semantic map for indefinite pronoun functions. Based on the map from \textcite[Figure 4.4]{Haspelmath1997}.}
\label{map:indef:haspelmath}
\end{figure}

As an example, Figure \ref{map:indef:haspelmath} displays the semantic map proposed by \citeauthor{Haspelmath1997} for the functions of indefinite markers across languages. The map shows, for example, that the functions of \textit{specific known} and \textit{specific unknown} are expressed by a single form in at least one language given the edge between these two nodes. No such relation exists between the functions of \textit{direct negation} and \textit{free choice}: if a lexical item exists that expresses these two functions, it should at least also express the functions of \textit{indirect negation} and \textit{comparative}. The distribution of language-specific items can then be mapped upon the semantic map in Figure \ref{map:indef:haspelmath} (cf.\ Figures 4.5--4.8 in \cite{Haspelmath1997}).

In current terminology, semantic maps as in Figure \ref{map:indef:haspelmath} are called \textit{classical maps}, while maps generated through MDS (or related methods) are generally called \textit{proximity maps} \parencite{VanderAuwera2013}.\footnote{See \textcite{Georgakopoulos2019} for a comprehensive overview of related terms.} Classical maps are lauded for their neutral theoretical stance and the falsifiable predictions that they can generate (i.e., possible and impossible polysemy patterns) and are widely used (see e.g.\ \cite{Georgakopoulos2018} and \cite{Georgakopoulos2019} for an overview of current research). Whereas classical maps are usually hand-crafted (but see \cite{Regier2013} for an algorithm), MDS maps aim to advance the semantic map method by generating maps directly from linguistic data, while at the same time upholding the neutral perspective. However, interpreting MDS maps is not trivial, as we will show in this paper.

Early work used MDS maps in an attempt to capture the insights of classical maps in a more systematic or automated way. More recently, researchers have started to employ MDS to map data from parallel corpora, intending to investigate a certain linguistic phenomenon from a cross-linguistic perspective without prescribing functional categories. In this paper, we focus on this latter trend in the use of MDS in linguistics, although the link between MDS and classical maps will be discussed in passing as we follow the chronological development of MDS maps in section \ref{sec:types}.\footnote{This paper is only concerned with MDS used as a method to create semantic maps. Other uses, such as the creation of areal maps in dialectometry \parencite{Wieling2015} or modeling diachronic change \parencite{Hilpert2011} are not discussed. Moreover, this paper does not aim to provide a practical user guide about software packages that implement MDS, see for instance \textcite{Croft2013} or \textcite[Appendix A]{Borg2005} for this purpose.}

We stress that MDS as a statistical technique does not stand for a single concept: MDS can be used to generate various kinds of maps, which show different things and have different functions in the context of linguistic argumentation. To explain what a given MDS map represents, we discuss different sorts of MDS maps based on three parameters: input data (what sort of linguistic data have been used as input for the MDS algorithm?), similarity measure (how is similarity between primitives defined?), and output (what do the data points on the map represent?). We advance these parameters as a concise way to provide the essential information for understanding how a given MDS map was constructed.

Before we proceed to discuss several MDS maps along these parameters, we first describe the mathematical foundations of MDS in section \ref{sec:math}. This exposition helps to understand the fundamental role of similarity in the construction of the maps, and familiarizes the reader with some essential terminology (\textit{eigenvalues}, \textit{stress factors}, \textit{dimensionality}) that is needed to understand the central concepts of multidimensional scaling. Then, in section \ref{sec:types}, we review various MDS maps that can be found in the literature since \textcite{Croft2008}. We look at the various types of linguistic input data, and explain how these MDS maps were constructed. Section \ref{sec:interpretation} covers how MDS maps can be interpreted by analyzing the dimensions of the map and the clustering of points, both by informal inspection and with the help of statistical tools. We also describe how this interpretation process links up to linguistic theory, by reviewing the types of research questions that MDS maps have been used to answer (section \ref{sec:theory}). In section \ref{sec:future}, we indicate promising future developments of MDS in the linguistic domain, as well as alternatives to MDS. Section \ref{sec:conclusion} concludes.

\section{MDS: Mathematical background} \label{sec:math}

This section gently introduces some of the mathematical concepts behind MDS, intended for readers who do not have a background in matrix algebra, but do want to understand notions used in the linguistic MDS literature such as `eigenvalues' and `stress factor'. Much more thorough expositions on the mathematics behind MDS are available, for example \textcite{Borg2005}. Readers who are primarily interested in linguistic applications of MDS may skip ahead, and continue reading at section \ref{sec:types}. Subsection \ref{sec:mathsummary} highlights the main points of section \ref{sec:math}.

Although in the linguistic literature the label `multidimensional scaling' is typically used without further qualification, MDS actually stands for a family of methods and procedures consisting of numerous variants that have been developed for different applications. Here, we introduce in some detail the version of MDS that is usually known as \textit{classic scaling} or \textit{classic MDS}, or more fully as \textit{classic metric Torgerson scaling}, named after the work of \textcite{Torgerson1952}. We opt for this variant for expository reasons -- it is the conceptually simplest model, and contains the core concepts needed to understand multidimensional scaling and its related technical concepts. 

Classic scaling is one of three MDS algorithms that have been used in linguistic applications of MDS, the other two being an iterative procedure known as SMACOF and an algorithm known as Optimal Classification (OC) MDS (see Supplementary Materials for a brief discussion of these). For brevity of reference, we will continue the terminological abuse by referring to classic scaling simply as `MDS' in this section.

The main component of MDS is a process called \textit{eigendecomposition}. This process is also used in other statistical techniques, such as \textit{Principal Component Analysis} \parencite{Jolliffe2016}. What is specific about MDS is that it uses as input for eigendecomposition a set of similarity or dissimilarity data between objects. We start our exposition at a general level and describe the mathematical principles underlying eigendecomposition (\S\ref{sec:eigen}), and then zoom in on some mathematical specifics of MDS, and the similarity data used as input (\S\ref{sec:similarity}).

\subsection{Matrix algebra and eigendecomposition}\label{sec:eigen}

MDS is based on matrix algebra. Matrices can be added and multiplied, just like numbers can. Matrix addition and the multiplication of a matrix by a number (also known as `scalar multiplication') are straightforward, as the following (arbitrary) examples illustrate:
\[ \begin{array}{l@{\hspace{1cm}}l}
\begin{pmatrix} 4 & 1 \\ -2 & 6  \end{pmatrix} + 
 \begin{pmatrix} -3 & 7 \\ 1 & -1  \end{pmatrix} = 
  \begin{pmatrix} 1 & 8 \\ -1 & 5  \end{pmatrix} & \text{[matrix addition]} \\
     3 \begin{pmatrix} 3 & -4 \\ 0 & 2  \end{pmatrix} =  \begin{pmatrix} 9 & -12 \\ 0 & 6  \end{pmatrix} 
     & \text{[scalar multiplication]}
     \end{array} \]
More important is how two matrices are multiplied. Matrix multiplication can be interpreted geometrically. This is easiest when we multiply a $n \times n$ matrix by a vector of length $n$. A vector is an arrow in $n$-dimensional space, so it has a length and a direction. It can be written as a matrix with $n$ rows and 1 column (or 1 row and $n$ columns). An example of matrix multiplication (with arbitrarily chosen numbers) is given below:\footnote{Matrix multiplication is only possible if the two matrices have suitable sizes: a $m \times n$ matrix can be multiplied with a $n \times k$ matrix to yield a $m \times k$ matrix. The values in the resulting matrix are determined by \textit{dot products} of rows in the first matrix, and columns in the second matrix (see any linear algebra textbook for definitions). In the example, the coordinates of the output vector are found by $-2 \times 4 + 2 \times -2 = -12$ and $-3 \times 4 + 5 \times -2 = -22$.}
\begin{equation}\label{eqn:mult}
 \begin{pmatrix} -2 & 2 \\ -3 & 5  \end{pmatrix} 
\begin{pmatrix} 4  \\ -2   \end{pmatrix} = 
\begin{pmatrix} -12 \\ -22   \end{pmatrix}
\end{equation}
Writing the matrix as $\mathbf{A}$, and the input and output vectors as $v$ and $w$, we can represent the above equation as $\mathbf{A} v = w$. We can understand the multiplication by $\mathbf{A}$ as a geometric transformation such as rotation, scaling, reflection, etc. In other words, the coordinates of $\mathbf{A}$ can be chosen in such a way that $\mathbf{A}$ acts like a geometric operator that maps an input vector $v$ to an output vector $w$. 

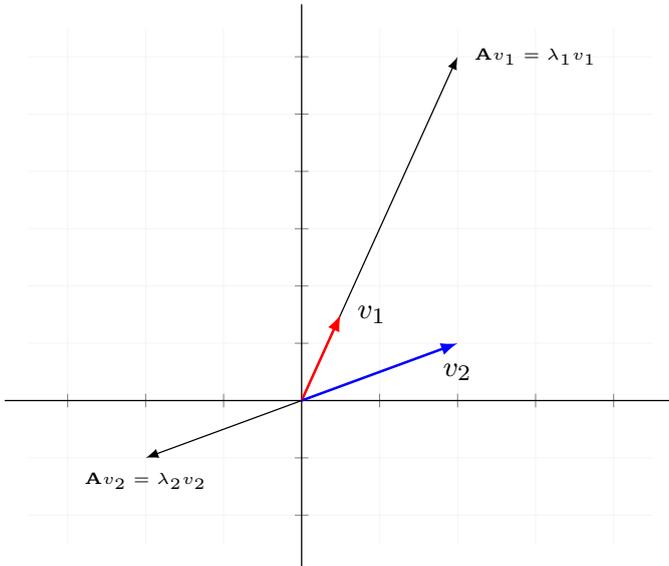
\begin{figure}[!ht] \centering
\begin{tikzpicture}[scale=1.2]
\begin{axis}[grid=both,grid style={line width=.1pt, draw=gray!10},major grid style={line width=.1pt,draw=gray!10}, clip=false,
    xmin=-3,xmax=4,
    ymin=-2,ymax=6,
    axis lines=middle,
    enlargelimits={abs=0.5},
    xtick={-7,-6,...,7},
    ytick={-5,-4,...,6},
    xticklabels={\empty},
    yticklabels={,,},
    yticklabel style={anchor=south east},
    axis line style={-},
    axis line style={shorten >=-7.5pt, shorten <=-7.5pt},
    xlabel style={at={(ticklabel* cs:1)},anchor=north west},
    ylabel style={at={(ticklabel* cs:1)},anchor=south west}
]

\addplot[-latex,samples=2,domain=0:2] {3 * x};
\addplot[-latex,samples=2,domain=0:.5,thick,red] {3 * x};
\addplot[-latex,samples=2,domain=0:2,thick,blue] {(1/2) * x};
\addplot[latex-,samples=2,domain=-2:0] {(1/2) * x};

\node at (axis cs:.9,1.5) {$v_1$};
\node at (axis cs:3,6) {\tiny $\mathbf{A} v_1 = \lambda_1 v_1$};
\node at (axis cs:2,.5) {$v_2$};
\node at (axis cs:-2,-1.4) {\tiny $\mathbf{A} v_2 = \lambda_2 v_2$};

\end{axis}
\end{tikzpicture}
\caption{Eigenvectors of $\mathbf{A}=\begin{pmatrix} -2 & 2 \\ -3 & 5  \end{pmatrix}$ with a positive eigenvalue $\lambda_1 =4$ and a negative eigenvalue $\lambda_2 = -1$.} \label{fig:eigen}
\end{figure}

A special case arises when $\mathbf{A} v = \lambda v$, i.e.\ the result of applying $\mathbf{A}$ to $v$ results in a vector with the same or opposite direction, only scaled by a factor $\lambda$ (every coordinate of $v$ is multiplied by the number $\lambda$). If this happens, $v$ is called an \textit{eigenvector} of $\mathbf{A}$, and $\lambda$ its corresponding \textit{eigenvalue}.\footnote{For traditional reasons eigenvalues are denoted by $\lambda$. This has nothing to do with the lambda-operator in semantics.} Matrix $\mathbf{A}$, as used in equation (\ref{eqn:mult}), has eigenvalues $\lambda_1 = 4$ and $\lambda_2 = -1$.\footnote{There is a computational procedure to find eigenvalues and eigenvectors for a given matrix. For reasons of space, we do not outline this procedure here, but this can be found in any textbook on linear algebra.} The corresponding eigenvectors $v_1 = (\frac{1}{2}, \frac{3}{2})$ ($\mathbf{A} v_1 = (2,6)$ has the same direction, but stretched by a factor 4) and $v_2 = (2,1)$ ($\mathbf{A} v_2 = (-2,-1)$ has opposite direction, and same length) are displayed in Figure \ref{fig:eigen}.\footnote{Note that there are infinitely many eigenvectors: all multiples of $(\frac{1}{2}, \frac{3}{2})$, such as $(1,3)$, $(2,6)$, $(100,300)$ etc.\ are eigenvectors for $\lambda_1 = 4$. What is relevant is the number of eigenvectors that are \textit{linearly independent} (cannot be written as combinations of scalar multiples of each other). For a $n \times n$ matrix, there are at most $n$ linearly independent eigenvectors.}

Eigenvectors and eigenvalues have many applications in mathematics and statistics. For our (linguistic) purposes, the main motivation for using them is that they can help reduce a complex dataset to one of lower dimensionality. Suppose we write the dataset as a matrix (for example individuals for rows, and observations for columns). Then the eigenvectors of that matrix can be informally thought of as the dimensions along which most variation in the dataset occurs.\footnote{This is a bit of a simplification: technically, the eigenvectors are computed for the \textit{covariance matrix}, in a process called Principal Component Analysis (PCA). For reasons of space we leave out visualizations of this, but see e.g.\ \textcite[2-3]{Jolliffe2002} for plots of principal components of a bivariate dataset.} The eigenvalue corresponding to an eigenvector indicates the relative significance of that eigenvector's dimension in describing the data.

Eigenvectors and eigenvalues have a further special property: for most matrices $\mathbf{A}$ -- and in particular \textit{symmetric} matrices,\footnote{A matrix is symmetric if the value $a_{i,j}$ at row $i$ and column $j$ is the same as the value $a_{j,i}$ at row $j$ and column $i$, for any $i$ and $j$. A real-valued symmetric matrix has the property that it always has real-valued eigenvalues and eigenvectors.} which will show up in the setting of MDS -- it is possible to reconstruct the matrix $\mathbf{A}$ by only using the eigenvectors/values. This process is called \textit{eigendecomposition}, which is to say that $\mathbf{A}$ can be written as a product of three matrices, as follows:
\[ \mathbf{A} = \mathbf{Q} \mathbf{\Lambda} \mathbf{Q}^{-1} \]
Here, $\mathbf{Q}$ contains the eigenvectors of $\mathbf{A}$ as its columns, and $\mathbf{\Lambda}$ (capital Greek letter lambda) is a diagonal matrix containing the eigenvalues of $\mathbf{A}$, which means that all its entries are 0 except for the ones on the diagonal, which contain the eigenvalues $\lambda_1, \ldots, \lambda_n$ of $\mathbf{A}$:
\[ \mathbf{\Lambda} = \begin{pmatrix} \lambda_1 & 0 & 0 & 0 \\
0 & \lambda_2 & \ddots  & 0 \\
0 & \ddots & \ddots & 0 \\
0 & 0 & 0 & \lambda_n
\end{pmatrix}
\]

$\mathbf{Q}^{-1}$ is the \textit{inverse} of $\mathbf{Q}$, which is to say that the product $\mathbf{Q} \mathbf{Q}^{-1}$ is the unit matrix, the diagonal matrix with ones on its diagonal.\footnote{The counterpart in the domain of numbers is that the `inverse' of the number $a$ is $\frac{1}{a}$, because the product $a \times \frac{1}{a}$ is the unit number 1.}

For our example matrix $\mathbf{A}$ from Figure \ref{fig:eigen}, the eigendecomposition is as follows:
\[ \mathbf{A} = \begin{pmatrix}
1 & 2 \\ 3 & 1 \end{pmatrix} \begin{pmatrix} 4 & 0 \\ 0 & -1 \end{pmatrix} \begin{pmatrix}
-\frac{1}{5} & \frac{2}{5} \\
\frac{3}{5} & -\frac{1}{5} 
\end{pmatrix}.\]
In general, applying eigendecomposition to a data matrix reveals the most important dimensions in the data (eigenvectors, from $\mathbf{Q}$), and the relative importance of those dimensions (eigenvalues, from $\mathbf{\Lambda}$). 

\subsection{Applying MDS} \label{sec:similarity}

The input for MDS are (dis)similarity data. Similarity between two objects $i$ and $j$ is represented as a numerical value $a_{i,j}$. Because the similarity between $i$ and $j$ is equal to the similarity between $j$ and $i$ ($a_{i,j} = a_{j,i}$), a (dis)similarity matrix is always a square symmetric matrix: 
    \[ \begin{pmatrix} 
    0       & a_{1,2} & a_{1,3} & \ldots & a_{1,n} \\
    a_{2,1} & 0       & a_{2,3} & \ldots & a_{2,n} \\
    a_{3,1} & a_{3,2} & 0       &        & \\
    \vdots  & \vdots  &         & \ddots & \\
    a_{n,1} & a_{n,2} &         &        & 0
    \end{pmatrix} \]
If the value $a_{i,j}$ increases as objects $i$ and $j$ become more similar, we speak of a similarity matrix. If the value decreases as the objects become more similar, we speak of a dissimilarity matrix. One can think of a matrix of driving distances between cities as a natural example of a dissimilarity matrix.

To apply MDS to linguistic data, these data must come in the form of a (dis)similarity matrix. It may at this point not be clear how linguistic data, such as translations or native speaker judgments, can be represented in such a way. Concrete examples of how linguistic data are turned into a similarity matrix are discussed in section \ref{sec:types}.

The steps in the classic scaling algorithm are as follows \parencite[\S12.1]{Borg2005}:
\begin{enumerate}
\item Start with a matrix of dissimilarities $\mathbf{\Delta}$.
\item Apply an operation of \textit{double centering} to the matrix of squared dissimilarities $\mathbf{\Delta}^2$. This does not affect the relative dissimilarities, but results in a matrix $\mathbf{B}$ in which the values are centered around the origin (the rows and columns add up to zero).
\item Eigendecompose $\mathbf{B}$ as $\mathbf{B} = \mathbf{Q} \mathbf{\Lambda} \mathbf{Q}'$.
 \item  Select the largest $n$ eigenvalues from $\mathbf{\Lambda}$. Each of them corresponds with a column in $\mathbf{Q}$. The coordinates of the points in the reduced $n$-dimensional space are then found by keeping the $n$ columns corresponding to the chosen eigenvalues, and removing the other columns.\footnote{For a small numerical example of this procedure, see \textcite[263]{Borg2005}.}
\end{enumerate}
Because the matrix $\mathbf{B}$ is always symmetric (the original $\mathbf{\Delta}$, being a dissimilarity matrix, was also symmetric), a mathematical result ensures that the eigendecomposition of $\mathbf{B}$ results in a matrix $\mathbf{Q}$ that is \textit{orthogonal}. This means that the inverse of $\mathbf{Q}$ is simply the \textit{transpose} of $\mathbf{Q}$ (obtained by turning the columns into rows), written here as $\mathbf{Q}'$.

\subsection{Stress and dimensionality selection} \label{sec:stress}

\textit{Stress} is another frequently used term in linguistic MDS literature. Stress measures the difference between the MDS output and the original dissimilarity data. As larger stress values indicate a worse fit, stress is a badness-of-fit measure. The most commonly used measure, \textit{Kruskal's stress}, is based on the sum of squared deviations between the original dissimilarity data and the found coordinates in the output representation.\footnote{There are various other stress measures -- see \textcite[\S11]{Borg2005} for full mathematical details of these and other fit measures, or \textcite{Ding2018} for a shorter exposition.}

Step four in the above procedure involves dimensionality selection. MDS is a dimensionality reduction technique, but the number of dimensions in the final MDS output is something the researcher chooses. Stress can be used to help determine the optimal dimensionality of the MDS output. One easy general procedure is to generate MDS outputs of increasing dimensionality $2, 3, \ldots, n$, and calculate the stress value corresponding to each one. By comparing these (decreasing) stress values, the `optimal' dimensionality $k$ can be determined as the one for which the stress value does not decrease much anymore for higher dimensions than $k$. This method is known as the `elbow method', after the shape of the line plot in a graphic representation of stress values for different dimensions (see \cite[Fig.\ 4]{Levshina2016LiC} for an example of such a plot). \textcite[\S3.5]{Borg2005} provide many more details on interpreting stress values.


Next to computing a global stress value, stress can be used to detect potential outliers in the dataset by focusing on the deviation of individual points, as in \textcite[15]{Levshina2020}. Hence, stress values are a more flexible method in analyzing the fit of an MDS output than eigenvalues (recall \S\ref{sec:eigen}), which are associated with a dimension as a whole.


\subsection{Short summary}\label{sec:mathsummary}
The main points of this section are:
\begin{itemize}
\item Multidimensional scaling (MDS) stands for a set of statistical tools that use matrix algebra to reduce a complex multidimensional dataset to a representation of lower dimensionality.
\item The basic algorithm of \textit{classic scaling} achieves this by using \textit{eigenvector} methods. In informal terms, eigenvectors represent the main axes of variation in the dataset. The structure of eigenvalues is used to determine the optimal number of dimensions in the solution.
\item The input data for an MDS analysis consist of a matrix of (dis)similarity values between (linguistic) objects. 
\end{itemize}
See the Supplementary Materials for some further reading suggestions.

\section{A typology of MDS maps} \label{sec:types}

We separate our discussion of MDS maps into two parts. This section is about the construction of the maps: which input data and what parameters have been used in generating the MDS map? In other words, we attempt to provide a typology of MDS maps. We postpone interpretation of MDS maps, and how that links up to linguistic theory, until section \ref{sec:interpretation}.
 
The discussion in this section is chronological, starting with a brief overview of MDS maps that aim to recreate classical maps (\S\ref{sec:recreate}), and a related type of MDS map in which the points represent sentence contexts (\S\ref{sec:contexts}). Then, we cover in more detail the recent trend of creating MDS maps based on parallel corpus data (\S\ref{sec:parallel} and \S\ref{sec:tmining}).
 
\subsection{Recreating classical maps} \label{sec:recreate}

The first type of MDS map is one that aims to recreate classical semantic maps. This was one of the early motivations of applying MDS in the linguistic domain: MDS was introduced because ``the semantic map model is in need of a sound mathematical basis'' \parencite[83]{Croft2007}. This was a methodological advancement, because MDS provided a way to automatize the process of building classical semantic maps, and made it possible to deal with large-scale sets of data that could not be analyzed manually. 

This type of MDS map is often based on questionnaire data: sentence contexts that have been selected or designed by the researcher to investigate a particular domain (e.g.\ the tense/aspect questionnaire in \cite{Dahl1985}, or the performative questionnaire in \cite{DeWit2018}). The questionnaire is applied by native speakers or fieldworkers in several languages, and the data obtained from these questionnaires serve as input for MDS. 

\begin{figure}[!ht]
     \centering
     \setkeys{draftfigure}{content={Figure 4 from \textcite{Croft2008}}}
    \includegraphics[width=8cm]{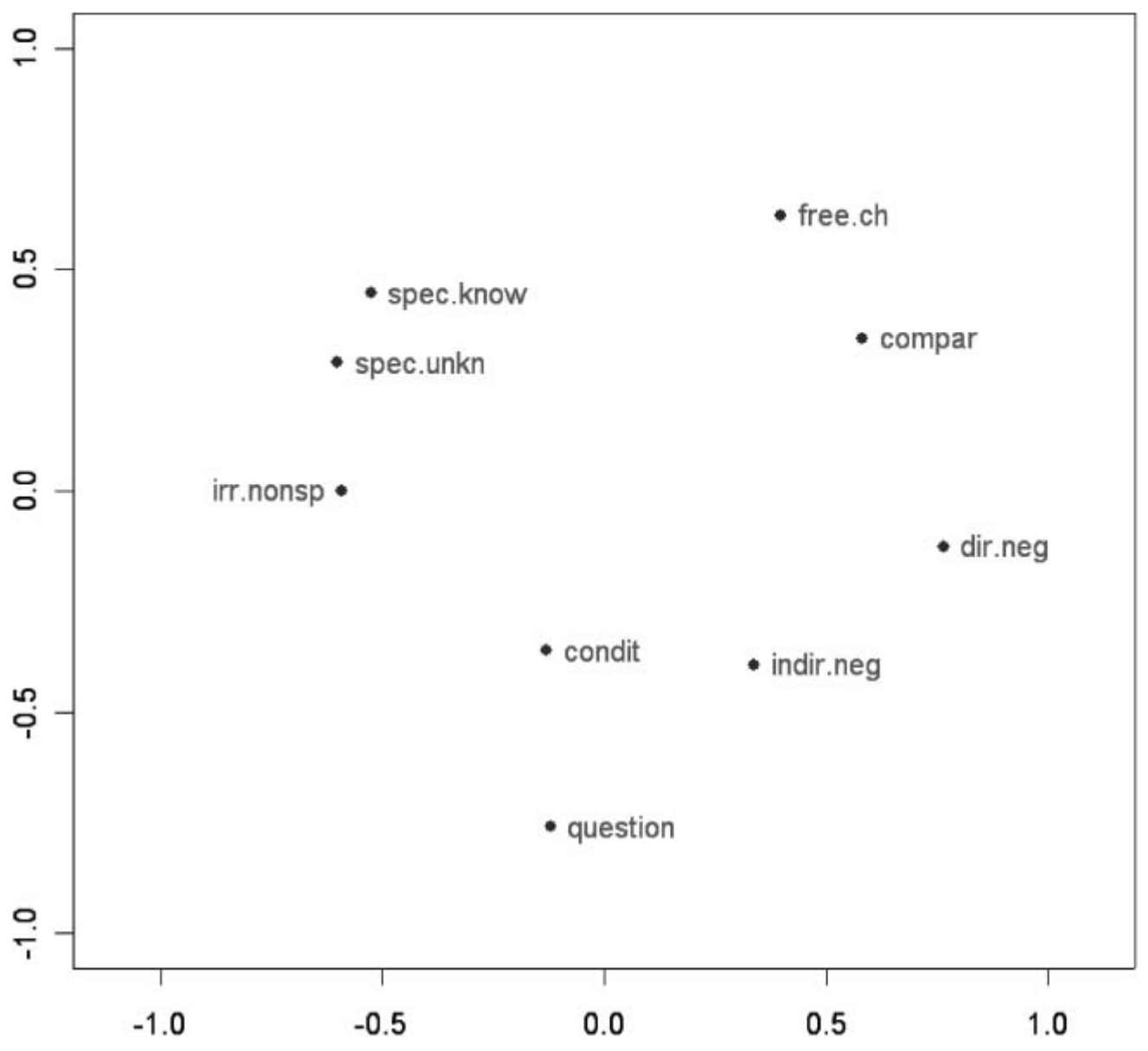}
    \caption{MDS map for indefinite pronoun functions. From \textcite[Figure 4]{Croft2008}.}
    \label{map:indef}
\end{figure}

In particular, the input data for these maps consist of specifications (Yes/No) for forms in various languages about whether or not that form can convey an abstract function. Two functions count as more similar when a higher number of forms express both functions. An example is Figure \ref{map:indef}, which displays an MDS map for indefinites from \textcite{Croft2008}, based on data from \textcite{Haspelmath1997}. The MDS map in Figure \ref{map:indef} reproduces the classical semantic map in Figure \ref{map:indef:haspelmath} (\S\ref{sec:intro}). The construction of the MDS map in Figure \ref{map:indef} is summarized in the box below it. We will use these boxes as a way to summarize the \textit{key parameters} of an MDS study, as given in section \ref{sec:intro} above: the algorithm used, the type of input data, how similarity was computed, and what the output map represents. The boxes use generic terminology such as `function$_i$', `construction$_j$', etc., to give the reader an understanding of how this type of MDS map works without specific details of any particular study. For the \textcite{Croft2008} study, the functions are the nine functions from \textcite{Haspelmath1997}, and the forms are indefinite pronouns from a variety of languages.

  \begin{tcolorbox}[title={MDS map based on \citeposs{Haspelmath1997} data on indefinites},breakable,enhanced jigsaw]
  \textbf{MDS algorithm:} Optimal Classification / unfolding
  
 \textbf{Input for MDS:} matrix of $n$ functions and $k$ forms, with Y(es) if that form conveys that function, and N(o) if it does not.
 \begin{center}
 \begin{tabular}{c|cccc}
 & function$_1$ & function$_2$ & \ldots & function$_n$ \\ \hline
   form$_1$ & Y & N & \ldots & N \\
   form$_2$ & N & Y& \ldots & N \\
   \vdots & \vdots & \vdots && \vdots \\
   form$_k$ & Y & Y& \ldots & N \\
 \end{tabular}
\end{center}
\textbf{Measure of similarity}: the similarity between two functions is measured by the number of forms that co-express them: d(function$_i$, function$_j$) = $\frac{\text{\#Y's in common}}{k}$. This way a $n \times n$ dissimilarity matrix is obtained.

 \textbf{Output map}: dots on the map represent abstract functions (the function$_i$'s), while distance on the map represents similarity between functions.
\end{tcolorbox}

Figure \ref{map:indef} aimed to recreate \citeposs{Haspelmath1997} classical semantic map of indefinites. Unlike in classical semantic maps, the distance between points is meaningful: points that are closer to each other are to be considered more similar.\footnote{A reviewer notes that also in work on classical semantic maps, extensions were developed that use distance between points as a way to display additional empirical generalizations. An example is \textcite{Nikitina2009}, who uses relative line length to represent differences in distance in conceptual space.} On the other hand, the dimensions have numerical values, but these do not have a direct linguistic interpretation. The dots on the MDS map may be connected to add the graph structure of the classical map (although this structure is not a result of the MDS algorithm), see \textcite[Fig.\ 6]{Croft2008}.

The similarity of this type of MDS maps to classical semantic maps entails that they are subject to some of the same shortcomings that classical maps have. For example, the literature on classical maps debates whether the abstract functions that are used as nodes in a classical map ought to be theory-neutral and comparable across languages, i.e.\ should be \textit{comparative concepts} \parencite{Haspelmath2003,Haspelmath2010}. It is not always easy to make sure that data satisfy this property, and this problem persists for MDS-based classical maps.

\begin{figure}[!ht]
    \centering
      \setkeys{draftfigure}{content={Figure 5 from \textcite{Croft2008}}}
    \includegraphics[width=8cm]{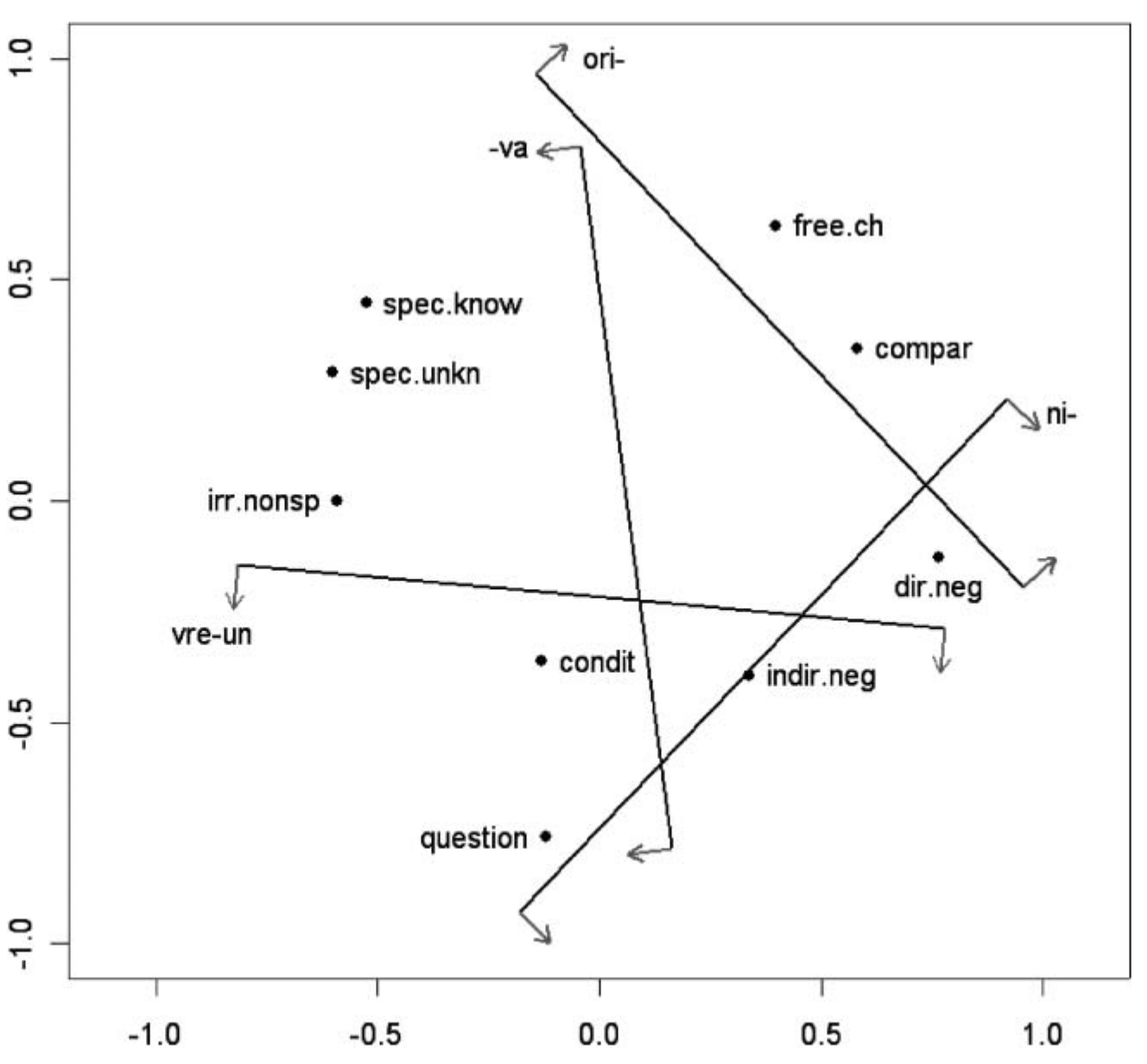}
    \caption{Figure \ref{map:indef} with cutting lines added for Romanian. From \textcite[Figure 5]{Croft2008}.}
    \label{map:cutting}
\end{figure}

Note that the points on the map in Figure \ref{map:indef} are multilingual abstractions, since they represent abstract functions that are positioned in the two-dimensional space based on how forms in various languages express these functions. However, a monolingual map can be created by adding \textit{cutting lines} to the map that indicate how language-specific forms realize the functions on the map. In Figure \ref{map:cutting}, this is illustrated for Romanian. For example, the cutting line that is labeled \textit{ori-} separates the functions (i.e., dots) on the map that the Romanian form \textit{ori-} `any' can convey (i.e.\ \textit{free choice} and \textit{comparative}) from functions that it cannot convey (for example \textit{specific known}, etc.). Cutting lines work in this setting because of the binary nature of the input data, but cannot be used for other types of MDS input data (we refer the reader to \cite{Poole2005} and \cite{Croft2008} for more details on cutting lines).

This way, this type of MDS maps allows for the same two perspectives as classical semantic maps do, as described in \textcite[9]{Georgakopoulos2018}: translational equivalents are visible in the MDS map as a whole, and designations of a particular meaning intra-linguistically appear in language-specific maps. 

Besides the work of \textcite{Croft2008}, other domains for which MDS maps of this type have been made include Slavic tense \parencite{Clancy2006}, person marking \parencite{Cysouw2007}, and causatives (\cite[\S2]{Levshina2020}). The latter study is noteworthy because it contains three-dimensional MDS maps that are construed based on data from language grammars \parencite[Figures 4 and 5]{Levshina2020}.

\subsection{Incorporating sentence contexts}\label{sec:contexts}

A variant of the type of MDS map described above appears in \citeposs{Croft2008} reanalysis of data from \textcite{Dahl1985}. While a map such as the one in Figure \ref{map:indef} is based on forms (indefinite pronouns) and abstract functions, it does not include the data on which it was decided that a certain form may express a certain function. These data typically come in the form of sentence contexts that purport to show that form $x$ can be used to express function $y$. \citeauthor{Croft2008}'s map of Dahl's data does include these underlying sentence contexts, but is otherwise conceptually similar to the maps discussed above in that it also involves an interpretation of the contexts in terms of abstract functions by the researcher.

\begin{figure}[!ht]
\centering
\setkeys{draftfigure}{content={Figure 8 from \textcite{Croft2008}}}
\includegraphics[width=10cm]{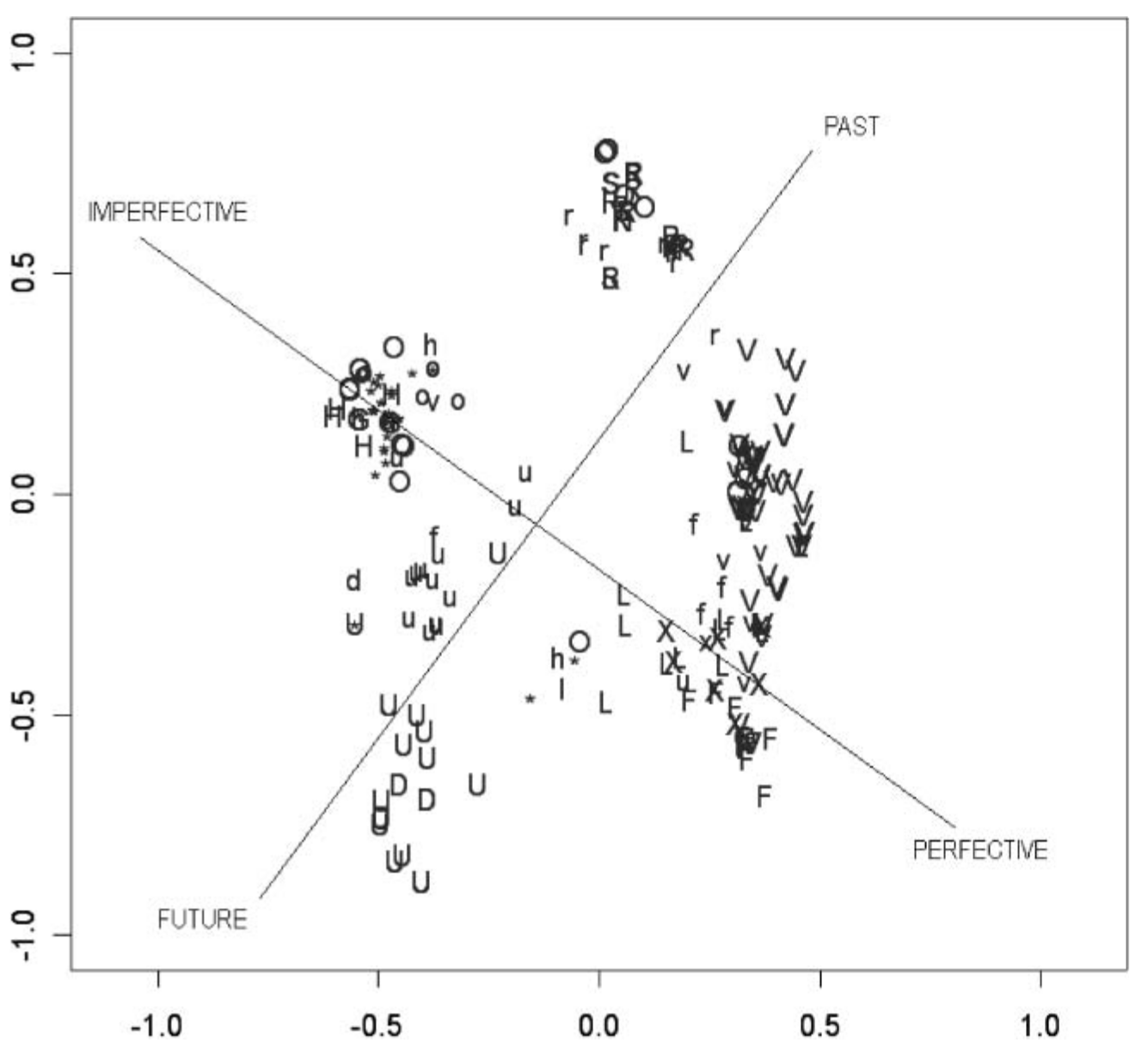}
\caption{MDS map of \citeposs{Dahl1985} tense-aspect data, with interpretative lines added. From \textcite[Figure 8]{Croft2008}.}
\label{map:dahl}
\end{figure}

The map, displayed in Figure \ref{map:dahl}, is based on \citeposs{Dahl1985} questionnaire on tense-aspect constructions in various languages. In this questionnaire, informants were asked to translate sentences in context (such as `He \textsc{write} a letter' in the context where you saw someone engaging in an activity yesterday, \cite[198]{Dahl1985}). The constructions cross-cut languages, and include for example `English \textit{simple present}', `French \textit{imparfait}', `Zulu \textit{narrative past}', etc. Croft and Poole assigned each of the 250 sentence contexts to a prototype (`perfective', `habitual', etc.). The contexts appear on the map as dots with a label for their prototype (such as the label V for `perfecti\ul{v}e'). As a result, a single label appears several times on the map. This type of MDS map is summarized in the box below, again presenting the input data in a generalized way.

Lastly, the lines on the map in Figure \ref{map:dahl} (past-future and imperfective-perfective) are added post hoc by Croft and Poole as an interpretation of the two dimensions of the MDS map. In section \ref{sec:dimension}, we return to the qualitative and quantitative assessment of the significance of MDS dimensions in more detail. 

  \begin{tcolorbox}[title={\citeauthor{Croft2008}'s map of Dahl's tense-aspect data}]
  \textbf{MDS algorithm:} Optimal Classification / unfolding
  
    \textbf{Input for MDS:} matrix of sentence contexts and constructions. 
    \begin{center}
   \begin{tabular}{l|cccc}
 &   sentence context$_1$ & sentence context$_2$ & \ldots & sentence cont.$_{250}$ \\
 & code: V & code: U & & code: r \\
 \hline
   construction$_1$ & Y & N & \ldots & N \\
   construction$_2$ & N & Y & \ldots & Y  \\
  \multicolumn{1}{c|}{\vdots} & \vdots & \vdots &&  \vdots \\ 
   construction$_{1107}$ & Y & Y & \ldots & N \\
   \end{tabular}
 \end{center}
 \textbf{Measure of similarity}: as above
 
 \textbf{Output map}: dots on the map are sentence contexts, represented by their prototype code
\end{tcolorbox}

MDS maps of a similar nature include the ones in \textcite{DeWit2018}, who use a questionnaire on aspectual constructions in performative contexts. \textcite{Hartmann2014} apply MDS to map microroles (verb-specific semantic roles) from 25 languages. Similarity between two microroles is based on co-expression tendencies between the two (see their p.\ 469 for details on the similarity measure).\\

\noindent \textbf{Map coloring}

\begin{figure}[!ht]
\centering
\setkeys{draftfigure}{content={Figure 5 from \textcite{Hartmann2014}}}
\includegraphics[width=.8\textwidth]{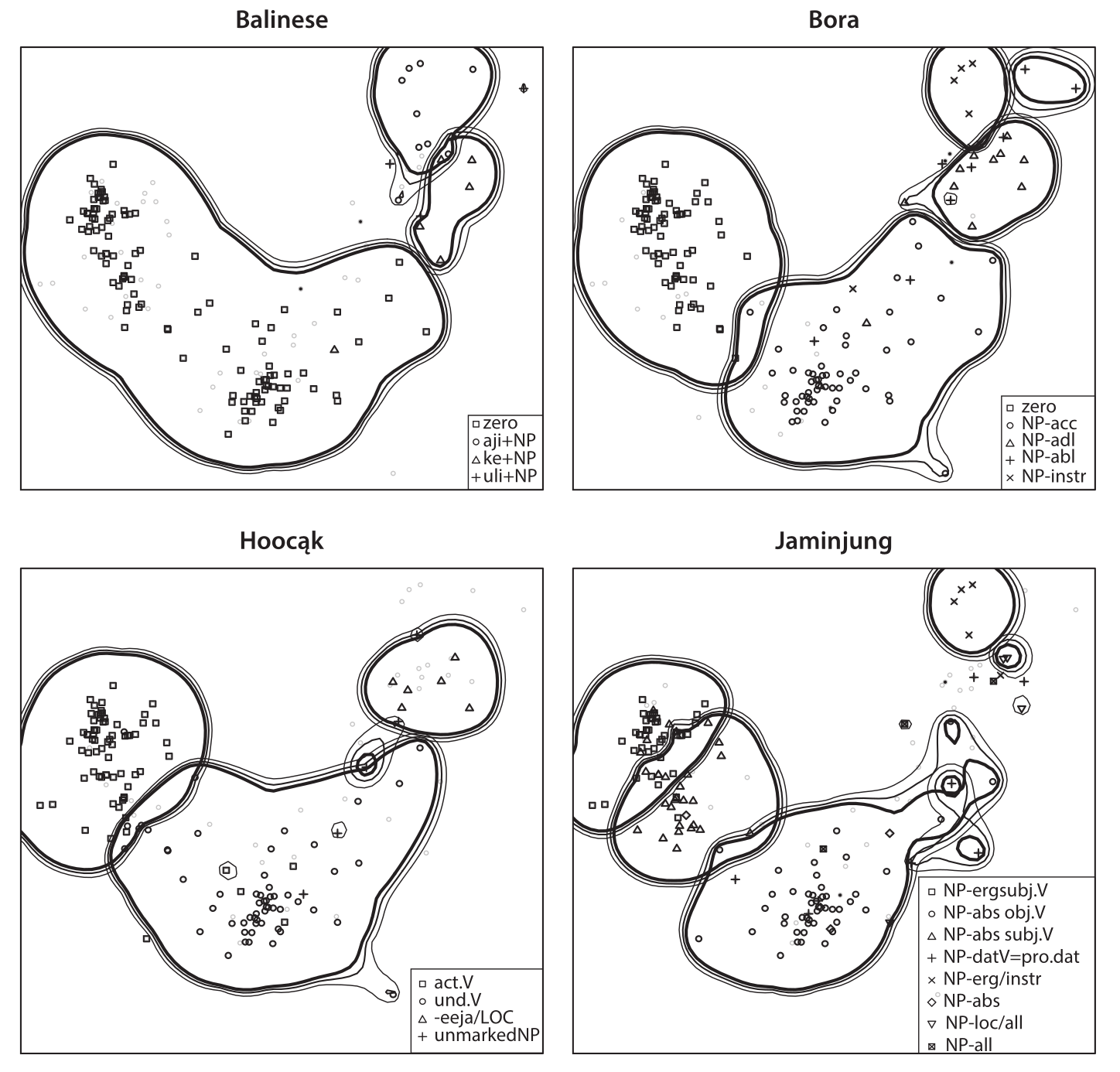}
\caption{MDS maps with different map coloring per language, with contour lines added. From \textcite[Figure 5]{Hartmann2014}.}
\label{map:hartmann}
\end{figure}

\noindent In the same way that cutting lines were used to display information about a specific language in a multilingual map (recall Figure \ref{map:cutting}, \S\ref{sec:recreate}), MDS maps that represent individual contexts can likewise display cross-linguistic variation. Language-specific constructions can be indicated by changing the appearance of the dots on the map (e.g.\ by using colors or symbols), a process we will refer to as \textit{map coloring}. Map coloring is used in many MDS studies (e.g.\ \cite{Walchli2010,Walchli2012}); here we illustrate with an example from \textcite{Hartmann2014}. Figure \ref{map:hartmann} shows the same map four times, but in each case the dots are represented differently, reflecting the constructions used in the four languages (the meaning of the contour lines on the map are discussed in section \ref{sec:cluster}).

Map coloring is in important technique in MDS maps, as it allows to see language-specific variation and cross-linguistic stability in the same visualization. We return to map coloring in the next sections for other types of MDS maps.

\subsection{Maps of parallel corpus data}\label{sec:parallel}

Besides questionnaire data, a second important source of data for linguistic MDS analyses is texts that have been translated in various languages, forming a \textit{parallel corpus}. \textcite[674]{Walchli2012} refer to this as \textit{primary data typology}, contrasting it with analyses based on higher-level sources such as reference grammars. Parallel corpora overcome some issues of data collection with classical maps: there is no dependency on existing comparative concepts, and using corpus data also allows to include frequency as a factor. On the other hand, it has been pointed out that a parallel corpus can be a limited source of data in that it may only provide a genre-specific perspective, might lack specific forms, and overuse prototypical forms \parencite{Levshina2020}.

Examples of parallel corpora that have been used in MDS analyses include Bible corpora \parencite{Walchli2010,Walchli2016,Walchli2018,Walchli2012}, translation corpora of novels \parencite{Verkerk2014,Klis2021}, Europarl (translated proceedings of the European parliament; \cite{Klis2017,Swart2021}), and a corpus of subtitles \parencite{Levshina2015, Levshina2020}. 

Once a suitable parallel corpus is selected, the construction of interest must be extracted and annotated. For example, \textcite{Walchli2012} extract 360 clauses describing motion events from translations of the Gospel of Mark in 101 languages (`doculects' in their terminology) (see \cite{Walchli2010} for a similar study with a different sample from the Gospel of Mark; see \cite{Walchli2016} for a study on perception verbs based on data from the Gospel of Mark).

Unlike the maps in section \ref{sec:contexts}, in the setting of parallel corpora, a context corresponds with a sequence of translations. A toy example would be $\langle$\textit{book}, \textit{libre}, \textit{Buch}$\rangle$ for the English, French, and German occurrences of that noun in a sentence from a parallel corpus. Similarity between contexts is then measured by a \textit{distance function} applied to two such sequences. Typically, the (relative) \textit{Hamming distance} is used as a distance function: a context is represented as a sequence of translations, and the distance between two sequences of $n$ objects is defined as the number of objects that differ (compared pointwise) divided by $n$. For example, the distance between $\langle A, B, C, D, E \rangle$ and $\langle A, B, X, D, Z \rangle$ is 2/5 because two of the five positions differ (the 3rd and the 5th).

Other distance functions are possible, such as the Levenshtein distance that has been used in several (non-MDS related) applications in linguistics (see e.g.\ \cite{Greenhill2011}). Another plausible option is to define a distance function \textit{ad hoc}, for example one that weighs certain components heavier than others, as in \textcite{Levshina2015} (see below for more details). However, we are unaware of work in the linguistic MDS literature exploring different distance functions and their effect on the resulting MDS output that leads to linguistic insights (but see section \ref{sec:composition}).

In general terms, the input data for this type of MDS are summarized in the box.\footnote{There are variants of this set-up. For example, \textcite{Levshina2016LiC} uses parallel corpus data, but with binary values (Yes/No). For a concrete implementation of her data matrix in Excel, see \textcite[Fig.\ 3]{Levshina2016LiC}.}
\begin{tcolorbox}[title={Parallel corpus MDS}]
    \textbf{Input for MDS:} matrix of languages and contexts. 
    \begin{center}
   \begin{tabular}{l|cccc}
 &   language$_1$ & language$_2$ & \ldots & language$_{n}$ \\
  \hline
   context$_1$ & translation$_{1,1}$ & translation$_{2,1}$ & \ldots &  translation$_{n,1}$\\
   context$_2$ &  &  & \ldots &   \\
  \multicolumn{1}{c|}{\vdots} & \vdots & \vdots &&  \vdots \\ 
   context$_{k}$ & translation$_{1,k}$ & & \ldots & translation$_{n,k}$  \\
   \end{tabular}
 \end{center}
 \textbf{Measure of similarity}: similarity between two contexts is determined by relative Hamming distance
 
 \textbf{Output map}: dots on the map are contexts
\end{tcolorbox}

There are several recent studies in which MDS has been applied to parallel corpus data. Here, we give a short overview of which kind of datasets have been used. In section \ref{sec:theory}, we return to most of these studies in more detail, to show how they use MDS maps in answering research questions in a variety of theoretical frameworks. 

\textcite{Walchli2018} investigates temporal adverbial clauses headed by words such as \textsc{until}, \textsc{before}, and \textsc{while}. Using a methodology similar to that of \textcite{Walchli2012}, he builds an MDS map representing contexts from the New Testament (NT) parallel corpus from 72 languages. 

\textcite{Verkerk2014} uses a parallel corpus built from translations of three different novels in 16 Indo-European languages to investigate the encoding of motion events. This results in a 3D MDS map, but instead of computing Hamming distance between contexts (as in \citeauthor{Walchli2012}'s case above), distances are computed between languages. Hence, the dots in \citeauthor{Verkerk2014}'s (\citedate[349]{Verkerk2014}) MDS map represent languages, and not individual contexts.

\textcite{Dahl2016} study perfects and the related category of iamitives (forms like English \textit{already}). They create an MDS map in which the points represent \textit{grams} (a word, suffix, or construction in a particular language with a specific meaning and/or function). They interpret the MDS space as a `grammatical space'. Using NT Bible translations from 1107 languages, the similarity between two grams (for example English \textit{Present Perfect} and Swahili \textit{-me-}) is determined based on how similar their distributions are across the text.

\textcite{Beekhuizen2017} study indefinite pronouns. Whereas \textcite{Haspelmath1997} uses data from grammars to build a classical semantic map, \citeauthor{Beekhuizen2017} use data from a parallel corpus of subtitles and an MDS analysis using the Optimal Classification algorithm (see section \ref{sec:recreate} above). They find a more fine-grained pattern by showing that some of Haspelmath's functions are infrequent, while a cluster analysis (see also \S\ref{sec:cluster}) finds a different grouping of semantic functions than in Haspelmath's map.

\Textcite{Swart2012} apply MDS to occurrences of two Greek prepositions, both of which encode source as their main meaning, based on a four-language sample of a parallel corpus of NT Gospels. The approach, including the similarity measure used, is similar to \textcite{Walchli2010}. They use a special variant of map coloring which they call ``semantic overlays'': they only display the points (i.e.\ occurrences of a preposition) that correspond with a given semantic role, such as elative, ablative, and partitive. This way they can interpret if the poles of a given dimension correspond to these semantic roles.

\textcite{Levshina2015,Levshina2016LiC,Levshina2020}, in a series of papers, applies MDS by stress majorization (see Supplementary Materials) in the domain of causatives. \textcite{Levshina2015} studies analytic causatives in 18 European languages with a constructed parallel corpus of film subtitles. The procedure is similar to that of \textcite{Walchli2012}, but the annotated features for each causative construction are assigned different weights \parencite[498]{Levshina2015}. \textcite{Levshina2016LiC} is a similar study with the same corpus, but focuses on verbs of letting (e.g. English \textit{let}, French \textit{laisser}) in 11 languages.

\subsection{Translation Mining} \label{sec:tmining}

\textcite{Klis2017} developed a variant of the basic methodology from \textcite{Walchli2012}, which they dub \textit{Translation Mining}. Instead of comparing translations by the lexical items that were chosen, they compare translations based on a grammatical feature, namely the tense form used. So, for \citeauthor{Walchli2012}, when comparing two constructions $w_1$ and $w_2$ in the same language, they count as equivalent if they are the same lexical item ($w_1 = w_2$). For \textcite{Klis2017}, on the other hand, $w_1$ and $w_2$ count as equivalent if they use the same tense form ($\text{Tense}(w_1) = \text{Tense}(w_2)$), but $w_1$ and $w_2$ need not be the same lexical item. In both cases, similarity of contexts is determined through the relative Hamming distance.\footnote{As a reviewer points out, similarly, \textcite{Waldenfels2014} compares grammatical labels rather than lexical forms. However, crucially, while \textcite{Klis2017} calculates distances between contexts, \textcite{Waldenfels2014} instead measure the dissimilarity between languages. As a result, he uses another paradigm for visualization, i.e. Neighbor-Nets. We briefly return to this method in section \ref{sec:cluster} below.}

A consequence of this methodological step is that after the relevant data are extracted from the parallel corpus, they also need to be annotated for the grammatical feature in question, the step of `tense attribution' in \textcite{Klis2017}. These authors have developed a software tool \textit{TimeAlign}\footnote{Source code for \textit{TimeAlign} is available via \url{https://github.com/UUDigitalHumanitieslab/timealign}.} to facilitate the process of annotation of parallel corpus data.

In an extension of the \citeyear{Klis2017} study, \textcite{Klis2021} investigate cross-linguistic variation of the \textsc{perfect} in West-European languages, where small caps indicate a cross-linguistic tense category comprising language-specific forms such as the English \textit{Present Perfect}, the French \textit{Pass\'e Compos\'e}, etc. (these tense categories are defined purely based on form, e.g.\ auxiliary+participle). The parallel corpus used in this work contains translations of the French novel \textit{L'\'Etranger} by Albert Camus (cf.\ \cite{Swart2007}), and the MDS maps are created by the SMACOF algorithm.

\begin{figure}[!ht]
\centering
\begin{subfigure}{.5\textwidth}
  \centering
  \includegraphics[width=.9\linewidth]{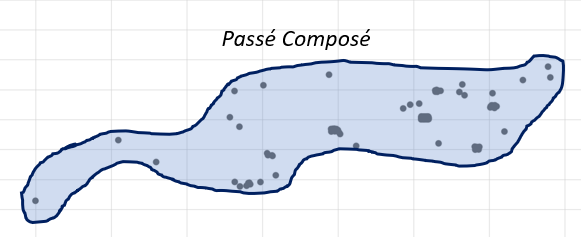}
  \caption{French}
  \label{fig:comp:fr}
\end{subfigure}%
\begin{subfigure}{.5\textwidth}
  \centering
  \includegraphics[width=.9\linewidth]{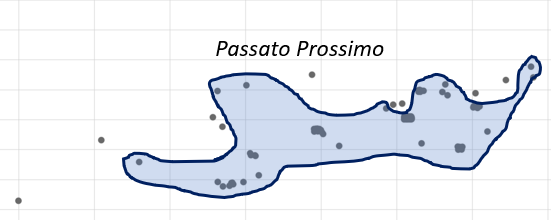}
  \caption{Italian}
  \label{fig:comp:it}
\end{subfigure}
\begin{subfigure}{.5\textwidth}
  \centering
  \includegraphics[width=.9\linewidth]{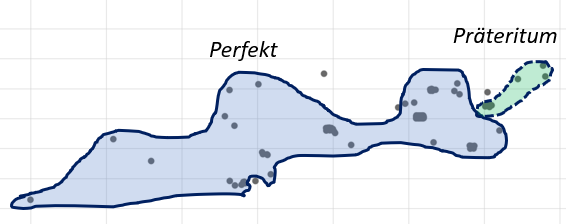}
  \caption{German}
  \label{fig:comp:de}
\end{subfigure}%
\begin{subfigure}{.5\textwidth}
  \centering
  \includegraphics[width=.9\linewidth]{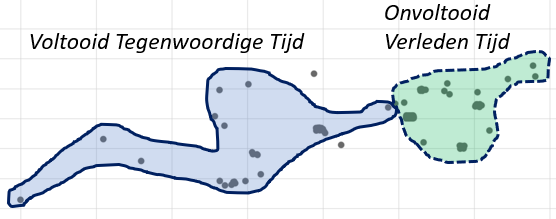}
  \caption{Dutch}
  \label{fig:comp:nl}
\end{subfigure}
\begin{subfigure}{.5\textwidth}
  \centering
  \includegraphics[width=.9\linewidth]{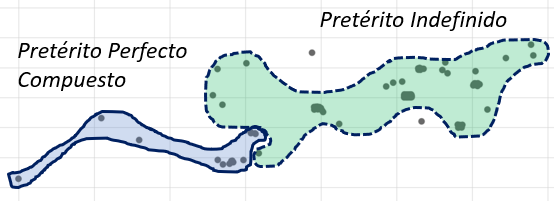}
  \caption{Spanish}
  \label{fig:comp:es}
\end{subfigure}%
\begin{subfigure}{.5\textwidth}
  \centering
  \includegraphics[width=.9\linewidth]{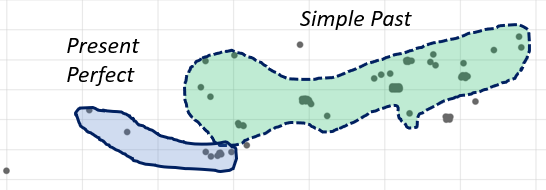}
  \caption{English}
  \label{fig:comp:en}
\end{subfigure}
\begin{subfigure}{.5\textwidth}
  \centering
  \includegraphics[width=.9\linewidth]{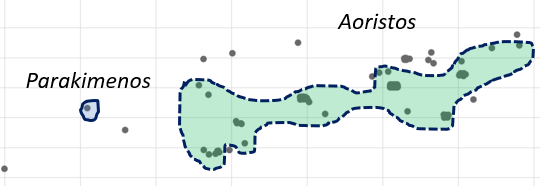}
  \caption{Greek}
  \label{fig:comp:el}
\end{subfigure}
\caption{MDS maps with different coloring per language, with added contour lines. The maps signal a subset relation between \textsc{perfect} and \textsc{past} in western European languages. From \textcite[Figure 3]{Klis2021}.}
\label{map:subset}
\end{figure}

A slightly different version of map coloring is used in this line of work: colors correspond to cross-linguistic tense categories, and not language-specific tense forms (so, for example, blue represents \textsc{perfect}). With this method, differences in tense use between languages can be identified. Figure \ref{map:subset} illustrates this: the same map is shown 7 times, but with colorings for the different languages in the corpus (blue for \textsc{perfect} and green for \textsc{past}). The stepwise reduction of the blue area (i.e.\ reduction of \textsc{perfect} use) is the visual representation of what \textcite{Klis2021} call a ``subset relation'' across western European languages' use of the \textsc{perfect}. There is a core use for which all languages use their counterpart of the \textsc{perfect} (blue), and then there is a scale from languages that use the \textsc{perfect} in only the core contexts (modern Greek) to languages that use it more widely (French, Italian). Further interpretation of the cut-off points between pairs of languages feeds a cross-linguistic semantic analysis of the \textsc{perfect}. Hence, MDS analysis is used to reveal a richer cross-linguistic variation in the domain of the \textsc{perfect} than was previously assumed in the literature (see \cite{Klis2021} for further details).

This study on the \textsc{perfect} gave rise to a line of (ongoing) work in which \textit{Translation Mining} is applied in other domains. \textcite{Bremmers2021} study definite determiners in German and Mandarin using a corpus of translations of \textit{Harry Potter and the Philosopher's Stone} by J.K.\ Rowling. \textcite{Tellings-ZRH} investigates variation in the domain of conditionals (see section \ref{sec:composition} below).\\

Having provided a typology of MDS maps in this section, in the next section, we turn to the interpretation of MDS maps. 

\section{Map interpretation and links to linguistic theory} \label{sec:interpretation}

Broadly speaking, there are two ways to analyze MDS maps. First, one can try to assign a linguistic interpretation to the dimensions of the map. We will call this process \textit{dimension interpretation}, and discuss this in \S\ref{sec:dimension}. Second, one can consider groups of points that cluster together on the map, a strategy that we refer to as \textit{cluster interpretation} (\S\ref{sec:cluster}). Note that dimension and cluster interpretation are not completely independent, as typically, when two clusters are separated on a map, they are also on opposing poles of one of the dimensions in the map. \S\ref{sec:theory} closes this section by linking interpretation of MDS maps to linguistic theory. We show that the MDS methodology is theory-neutral and has been used with different theoretical approaches, including classical typology and formal linguistics.

\subsection{Dimension interpretation}\label{sec:dimension}

Recall that the dimensions in an MDS solution do not have an intrinsic linguistic meaning, but are the outcome of the algorithm.\footnote{This relates to a general problem of visualizations that MDS maps are also subject to: they always show some structure in the data, even if this structure is only an artefact of the method applied \parencite[50]{Cysouw2008}. In this light, we should also view \citeposs{Zwarts2008} comment that the resulting dimensions in MDS maps do not necessarily reflect semantic distinctions.} Still, a typical desideratum of MDS studies is to interpret the dimensions so that the study assesses the distribution of points on the map qualitatively. For example, in Figure \ref{map:dahl} (\S\ref{sec:contexts}) the two dimensions are interpreted as a past-future axis and an imperfective-perfective axis. According to \textcite{Croft2008}, the first dimension displays cross-linguistic variation in tense: we find sentence contexts expressing past reference on the right side of the map, contexts expressing future reference on the left side, and finally, contexts that are generally not marked by grammatical tense (e.g., those expressing habituality) in the middle. The second dimension expresses aspect and has characteristically imperfective and perfective contexts on the extremes of the axis.

\begin{figure}[!ht]
     \begin{subfigure}{.48\textwidth}
     \centering
     \setkeys{draftfigure}{content={Figure 3 from \textcite{Walchli2012}}}
    \includegraphics[width=\linewidth]{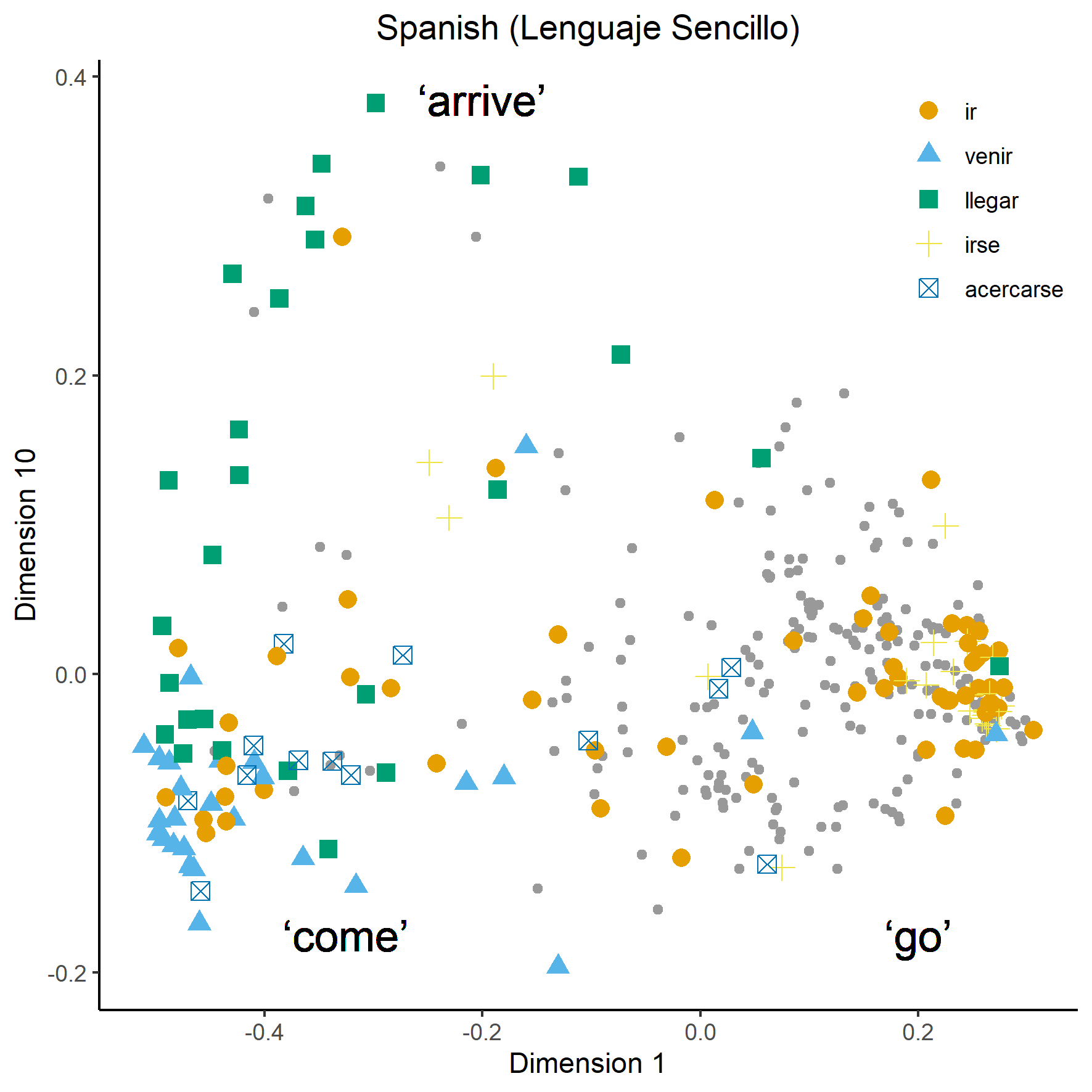}
    \caption{Spanish coloring} \label{map:WCspanish}
    \end{subfigure}
    \begin{subfigure}{.48\textwidth}
    \centering
    \setkeys{draftfigure}{content={Figure 3 from \textcite{Walchli2012}}}
    \includegraphics[width=\linewidth]{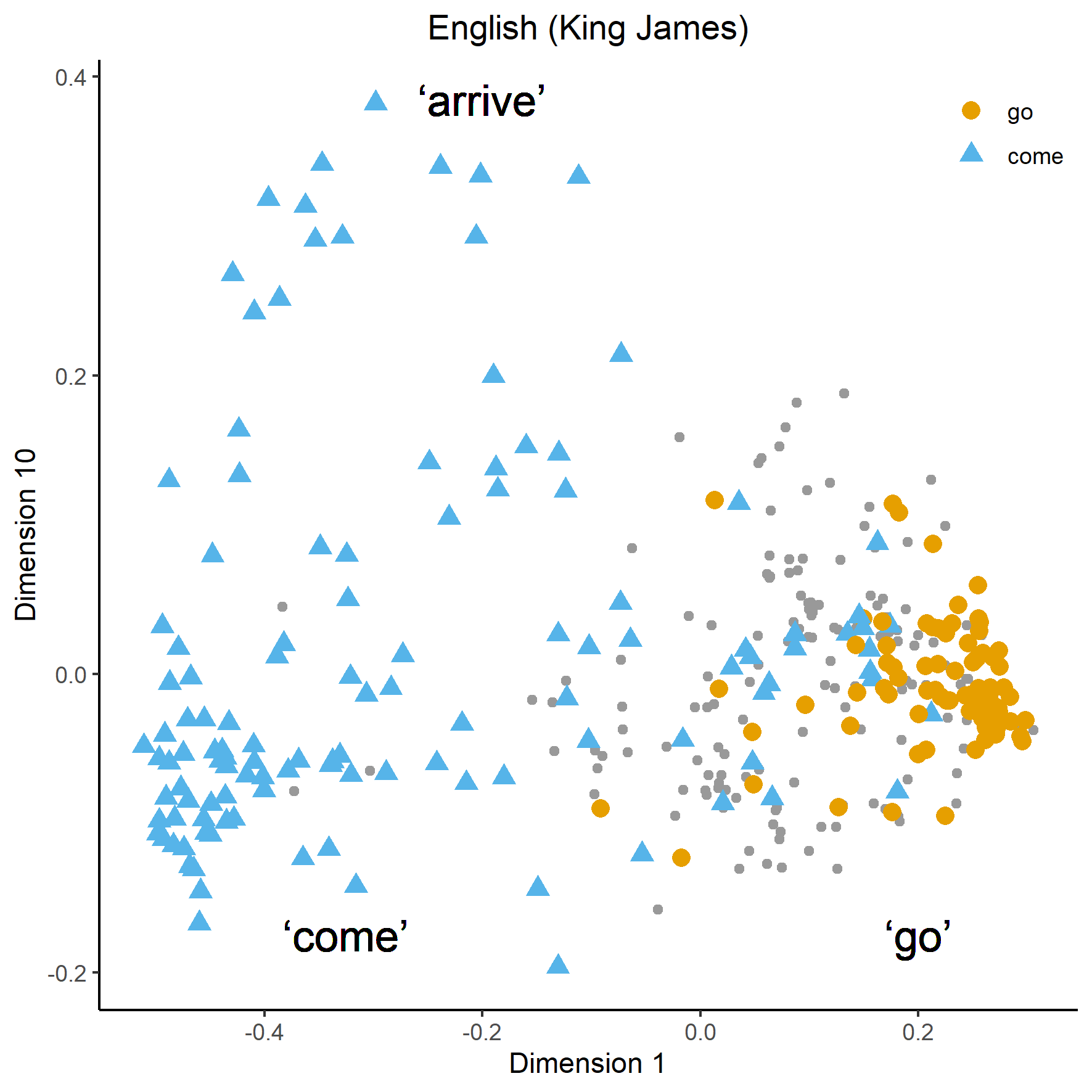}
    \caption{English coloring}\label{map:WCenglish}
    \end{subfigure}
    \caption{MDS maps with different coloring per language, with interpretative labels added. Based on the data from \textcite{Walchli2012}.}
    \label{map:WC}
\end{figure}

As another example, \textcite{Walchli2012} use eigenvalue analysis to find that at least 30 dimensions are relevant to describe their motion verb data. This number is rather high for linguistic MDS studies, and is taken by the authors to be illustrative of the high degree of complexity of the variation in the domain of motion verbs (p.\ 689). Instead of assigning a single interpretative label to each dimension, the authors separately interpret the negative and positive `pole' of a dimension. For example, dimension 1, having the highest eigenvalue and thus relatively the most important one (see \S\ref{sec:eigen}), is analyzed as distinguishing \textit{come/arrive} contexts (negative pole) from \textit{go/depart} contexts (positive pole, see their Table 4). As an example of how 2D maps are created for a high-dimensional MDS analysis, Figure \ref{map:WC} shows 2D maps plotting dimension 1 (\textit{come} vs.\ \textit{go}) on the x-axis and dimension 10, which distinguishes \textit{arrive} contexts at the positive pole, on the y-axis. This particular selection of dimensions allows \citeauthor{Walchli2012} to probe the cross-linguistic lexical variation in \textit{come}, \textit{go}, and \textit{arrive} contexts. As before, Figure \ref{map:WC} applies map coloring to indicate language-specific patterns on the map (Figures \ref{map:WCspanish} and \ref{map:WCenglish} display the same distribution of dots, but the coloring reflects Spanish and English, respectively). Labels are displayed in regions of the map corresponding with the poles of dimension 1.\footnote{The labels in Figure \ref{map:WC} correspond to \textit{regions} on the map, not clusters. See \textcite[690]{Walchli2012} for details on this rather subtle distinction.} 

One issue with the interpretation of dimensions is the potential occurrence of horseshoe patterns. For example, in Figure \ref{map:indef} (\S\ref{sec:recreate}), we find a pattern in which the functions \textit{specific known} and \textit{free choice} form two ends of a horseshoe. No cutting line in any language (cf.\ Figure \ref{map:cutting}, \S\ref{sec:recreate}) includes these two ends \parencite[18]{Croft2008}. As a consequence, one should interpret the functions as displaying only one dimension of variation, and not try to interpret the contribution of the x- and y-axis individually. Such a one-dimensional model here actually corresponds neatly with the hand-crafted classical map in Figure \ref{map:indef:haspelmath} (\S\ref{sec:intro}). 

Dimension interpretation often proceeds through visual inspection of MDS maps, but more rigorous approaches using statistical tools have also been proposed. \textcite{Levshina2020} uses linear regression to identify which of the semantic variables most strongly correlate with the placement of contexts in the MDS map (see also \cite{Levshina2011}). The procedure annotates the individual contexts of the MDS map with binary classifications (e.g.\ in the domain of causatives, one could annotate for contexts being \textit{intentional or not}, or \textit{factitive or permissive}). Regression analysis then correlates these variables with the positioning of a context on a single dimension. In other words, the method indicates which semantic phenomena best explain the cross-linguistic variation modeled by the MDS map.  

(Multiple) Correspondence Analysis is a method related to MDS, and facilitates dimension interpretation through the addition of supplementary points on the map. In the Supplementary Materials, we briefly introduce Correspondence Analysis. 

In the next section, we move from the interpretation of individual dimensions to the interpretation of clusters of data points on the MDS map. 

\subsection{Cluster interpretation and cluster analysis}\label{sec:cluster}

Groups of points that appear clustered on an MDS map are analytically relevant, because the proximity of the points indicates that the corresponding contexts are similar in a linguistically relevant way, and contrast with points outside the cluster. Clusters can be identified either by informal inspection of the map, or with the help of statistical or algorithmic tools. For example, the contour lines in Figure \ref{map:hartmann} (\S\ref{sec:contexts}) are obtained from a probabilistic method, see \textcite[471ff.]{Hartmann2014} for details. Once the clusters are identified, cluster interpretation is the process of inspecting the contexts from the dataset corresponding to the points in the cluster, and finding some linguistic commonality between them. For example, \textcite[470]{Hartmann2014}, in their MDS map of semantic roles, recognize clusters of agent-like roles and patient-like roles.

The procedure above consists of cluster identification and interpretation \textit{after} MDS has been applied to the dataset. An alternative is to identify clusters directly from the original dataset, and run MDS \textit{parallel} to it. Direct identification of clusters from the distance matrix (or a transformation thereof) is known as \textit{cluster analysis}. The resulting attribution of clusters to individual points can then be fed back to the MDS map as an additional layer of labelling. This procedure potentially facilitates the interpretation of the semantic dimensions at stake. Below, we describe two forms of cluster analysis that have been applied in combination with MDS. 

\subsubsection{\texorpdfstring{$k$}{k}-means clustering} \label{sec:kclustering}

$k$-means clustering aims to partition observations into $k$ clusters in which each observation belongs to the cluster with the nearest mean serving as a prototype of the cluster. $k$-means clustering can be applied to a distance matrix to find $k$ clusters consisting of similar data points. $k$-medoids clustering is a special case in which the center of each cluster is an actual data point; in $k$-means clustering, this need not necessarily be so. 

\begin{figure}[!ht]
\centering
\setkeys{draftfigure}{content={Figure 2 from \textcite{Walchli2018}}}
\includegraphics[width=.8\textwidth]{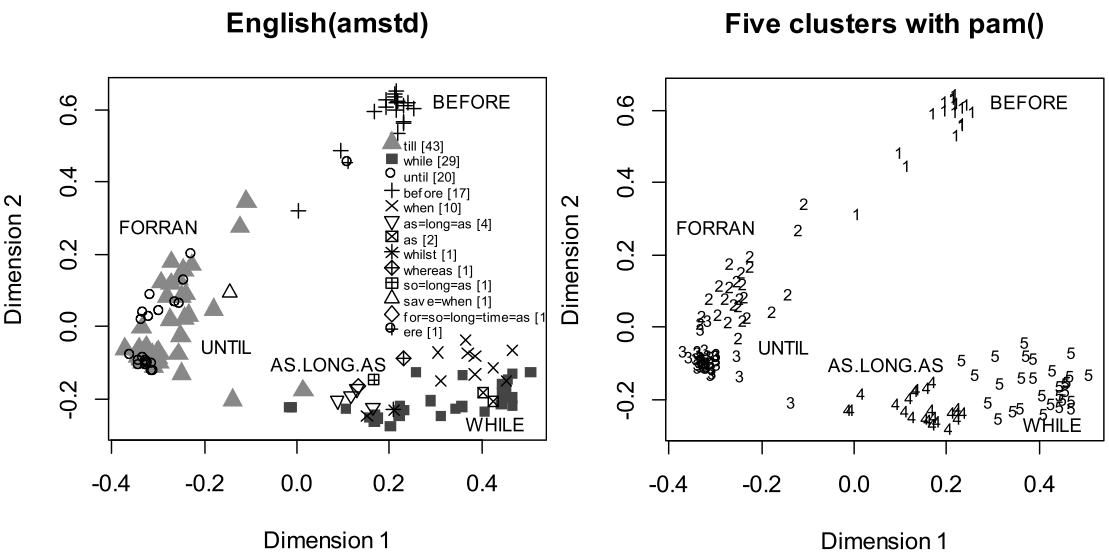}
\caption{On the left: MDS map with coloring for English, with cluster analysis through the Partitioning Around Medoids algorithm added. On the right: assignment of clusters to individual contexts by the Partitioning Around Medoids algorithm with $k=5$. Adapted from Figure 2 in \textcite[157]{Walchli2018}.}.
\label{map:pam}
\end{figure}

In \textcite{Walchli2018}, $k$-medoids clustering (in particular, the Partitioning Around Medoids algorithm) is applied to cross-linguistic lexical variation in the expression of adverbial clauses. With $k$ set to $3$, \textsc{as.long.as}, \textsc{until}, and \textsc{before} appear as three different semantic clusters.\footnote{Like in \textcite{Klis2021} mentioned above, small caps indicate a cross-linguistic category.} This result confirms earlier typological analyses in this domain, but without taking these functions as a point of departure, but rather as a result of cross-linguistic lexical variation. With $k = 5$, two additional clusters appear: \textsc{while} and \textsc{förrän} (from Modern Swedish \textit{förrän}, that is somewhere between \textit{before} and \textit{until}). Figure \ref{map:pam} shows MDS maps with additional labels for the identified clusters.

A post hoc analysis reveals that the optimal solution is with three clusters, and thus disregards \textsc{while} and \textsc{förrän} as meaningful clusters. From this result, one can infer that there are very few languages that have a separate lexical entry for \textsc{förrän} as Modern Swedish does. Instead, languages in general have the same marker for \textsc{förrän} and \textsc{until}. For English, the MDS map shows that there is a homogeneous distribution of \textit{till} and \textit{until} in these two clusters. A similar point can be made for \textsc{while}, that has a separate lexical marker in English, but which is cross-linguistically usually expressed with the same marker that expresses \textsc{as.long.as}.

\subsubsection{Hierarchical cluster analysis}

Hierarchical cluster analysis aims to build a hierarchy of clusters. The default, agglomerative variant takes a bottom-up approach: each observation starts in its own cluster, and pairs of clusters are iteratively merged while minimizing distance. The result is usually represented as a dendrogram. 

In \textcite{Levshina2020}, this type of cluster analysis is used to identify the semantic functions of causative constructions. Levshina annotated a typologically diverse sample of corpus subtitles and molded the parallel corpus data into the data structure posed in section \ref{sec:recreate} above. Hierarchical cluster analysis, as shown in Figure \ref{map:hca} below, then allows her to find seven clusters, that serve as the input for a semantic map. Using \citeposs{Regier2013} method to induce edges, Levshina ends up with a fully data-driven classical semantic map. 

\begin{figure}[!ht]
\centering
\setkeys{draftfigure}{content={Figure 1 from \textcite{Levshina2020}}}
\includegraphics[width=.8\textwidth]{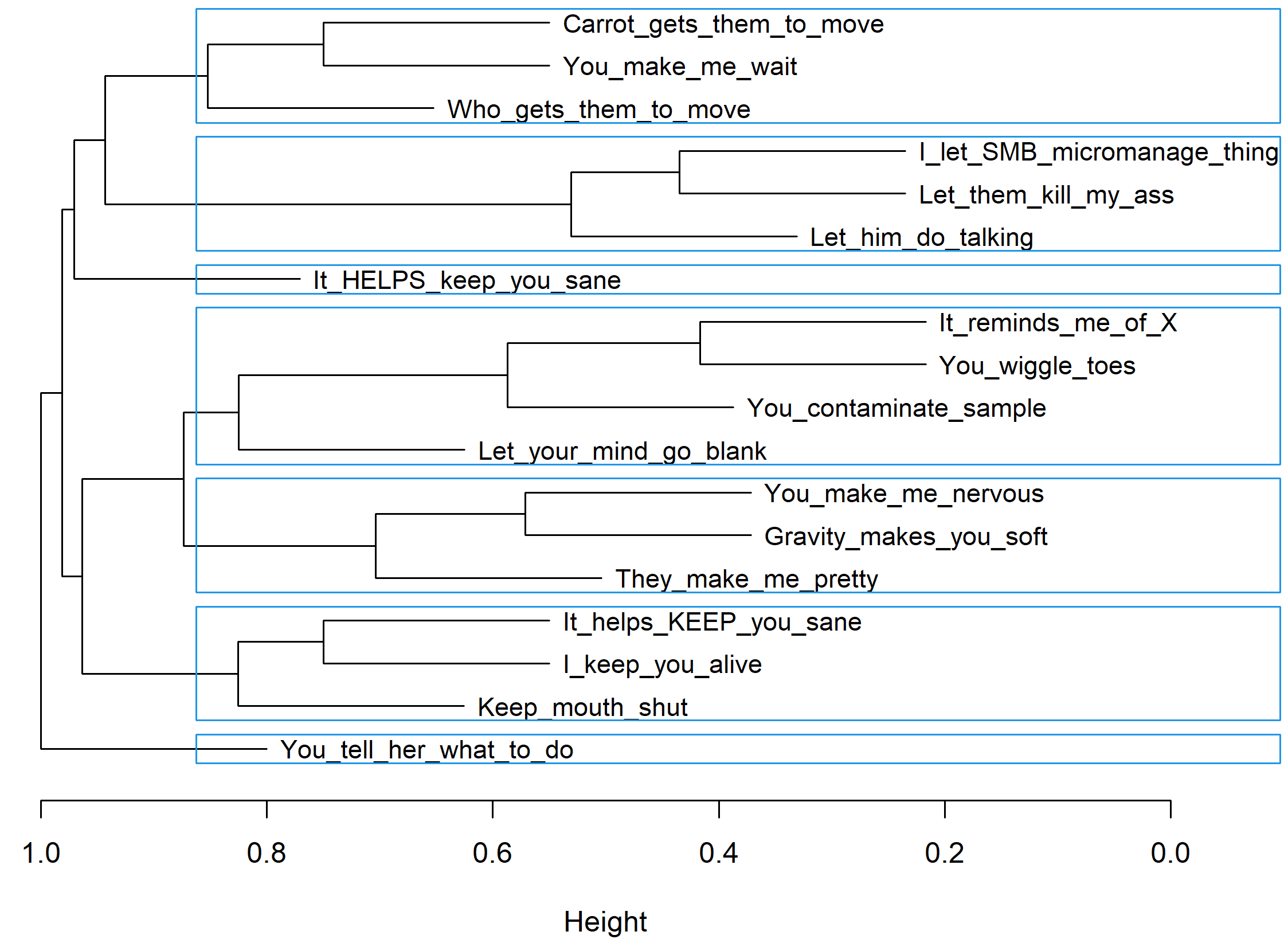}
\caption{Hierarchical cluster analysis on 18 causation contexts. The blue rectangles delimit the seven identified clusters. Based on the data from \textcite{Levshina2020}.}
\label{map:hca}
\end{figure}

Alternatively, not individual constructions, but rather languages as a whole are used as starting nodes of the hierarchical cluster analysis (e.g.\ in \cite[475]{Hartmann2014} and \cite[106]{Levshina2016LiC}). This move allows to generate hypotheses about genealogy or language contact, but crucially loses the possibility to drill down to individual contexts. Recently, Neighbor-Nets has been put forward as a related method that also operates on the language level and has similar aims \parencite{Bryant2004}, and has been successfully applied to parallel corpus data (e.g.\ in \cite{Dahl2014,Waldenfels2014,Verkerk2014,Verkerk2017}). \\

Cluster analysis and dimension analysis are interpretation methods for the map itself, but MDS studies in linguistics aim to answer some larger questions relating to linguistic theory. We now move to describe which part MDS maps play in the process of linguistic argumentation.

\subsection{MDS and linguistic theory}\label{sec:theory}

In this section, we discuss how multidimensional scaling as a data reduction and visualization technique stands in relation to theoretical approaches to the study of language. \textcite[8]{Georgakopoulos2018} point out that the semantic map method (in the broad sense as used in that work) is theory-neutral, and that this is one of its advantages: MDS can be used in combination with a wide range of descriptive and theoretical approaches of grammar, including formal and cognitive ones. We argue here that, likewise, the methodology of using parallel corpus data with an MDS analysis is theory-neutral. We illustrate this point by examining the studies cited in section \ref{sec:types} again, this time highlighting the theoretical contribution the authors aimed for by using MDS.

To illustrate the methodology's compatibility with a variety of theoretical approaches, we zoom in on two approaches in particular, `classic typology' and `formal linguistics' (to be defined below).\footnote{The names of the two approaches are put in scare quotes to indicate that the labels neither do justice to the large number of underlying assumptions that both approaches have (see e.g.\ \cite{Hawkins1988}), nor to the various variant and intermediate positions that exist.} We choose these for two reasons, first because most of the MDS studies we review can be positioned on a continuum between classic typology and formal linguistics (but this does not mean that we claim that no other frameworks are compatible with MDS). Second, the two approaches are sometimes perceived as contrastive or incompatible. For example, \textcite[85]{Croft2007} writes that ``typology starts with crosslinguistic comparison, while the structuralist/generative [i.e.\ formal] approach proceeds `one language at a time' ''. Our review will conclude that there is in fact no conflict, and that the MDS methodology adds a multi-language empirical basis to formal studies of linguistic phenomena.

\subsubsection{MDS as a theory-neutral method}

We will adopt the following idealized definitions of the two approaches. (Classical) typology is a form of inquiry in which large-sample linguistic comparison is applied to reveal limits of cross-linguistic variation in the form of (implicational, restricted, biconditional, \ldots) universals of language. Formal linguistics is an approach that, based on data from a single or a small number of languages, provides an in-depth abstract analysis of a given phenomenon that leads to an account that is deductive in the sense that it makes falsifiable predictions. We do not aim to review the debate here of how these two approaches relate to each other, and to what extent there is a conflict between them (see e.g.\ \cite{Croft2007,Cinque2007,Haspelmath2010,Newmeyer2010} for differing opinions).

Several studies are primarily interested in research questions about language classification, illustrating applications in classical typology. \textcite{Verkerk2014} is a clear example of this, whose aim is to check the validity of the ``strict dichotomy between satellite-framed and verb-framed languages'' (p.\ 326) proposed by \textcite{Talmy2000}. Her MDS maps are unusual in comparison to the studies discussed above in that they locate languages rather than semantic functions or linguistic contexts. From her MDS analysis, she concludes that a strict dichotomy cannot predict the attested variation, which gives rise to the potential identification of new language classes \parencite[351]{Verkerk2014}.

\textcite{Dahl2016} is an example of a large-sample MDS study (1107 languages). It addresses the question if two grams, perfects and iamitives, form two distinct clusters, or rather a continuum. The conclusion is that although certain areal groups can be identified as clusters in the MDS map, the distribution of grams forms a continuum.

\textcite{Hartmann2014} investigate the clustering of semantic microroles in a classic scaling MDS map. Through this map, a metric is computed that classifies languages based on pairwise similarity of microrole coding strategy. By this means, a hierarchical typology is constructed of the 25 languages in the study.

More towards formal linguistics is \textcite{DeWit2018}, who aim to investigate aspectual properties of performatives. They argue that, cross-linguistically, languages use the same aspectual category for performatives as they do for other constructions that have a similar epistemic property (see their \S2 for details). They use an MDS study to show that aspectual categories indeed pattern this way. This study can thus be argued to occupy somewhat of a middle ground: it is a typological study that investigates cross-linguistic patterns, but also aims to identify epistemic properties of performative and other speech acts.

In a similar position is \textcite{Walchli2012}, who employ MDS maps to represent the extent of variation in the domain of motion verbs (101 languages). Besides various methodological points, the authors apply detailed dimension and cluster interpretation on their MDS map to make typological and language-specific claims about the cross-linguistic variation of motion verbs. By inspecting the linguistic contexts behind the motion verbs, the authors propose a new category type `narrative \textit{come}' (p.\ 696), showing that the distribution of motion verbs also has a discourse component.

The study by \textcite{Klis2021} discussed in section \ref{sec:tmining} looks at a much smaller sample (seven European languages). However, this sample is sufficient to identify a subset relation in the use of the \textsc{perfect}, rather than a hitherto assumed dichotomy between strict and liberal \textsc{perfect} languages. This observation forms the starting point for a formal linguistic analysis of the contexts in which pairs of languages differ with respect to \textsc{perfect} use. 

\Textcite{Swart2012} represents a more radical departure from the typological studies discussed above in that it is primarily interested in a phenomenon in a single language -- the semantics of the source prepositions \begin{greek}>ap'o\end{greek} (\textit{apo}) and \begin{greek}>ek\end{greek} (\textit{ek}) in Ancient Greek. The authors use a parallel corpus MDS study to measure the semantic similarity between the two prepositions, stating explicitly that they want to investigate how the (broad-sample) MDS methodology ``can be applied to a small language sample'' (p.\ 163). By an analysis of the semantic features of the clusters on the map, they come to a better understanding of the semantic role of both prepositions.

Similarly, \textcite{Bremmers2021} are primarily interested in a phenomenon in a single language: how is the formal distinction between weak and strong definites operational in Mandarin? A small-sample MDS study, with only three languages (English, German, and Mandarin Chinese), shows that, contrary to earlier predictions, Mandarin bare nominals and demonstratives do not map directly on German contracted (weak definites) and uncontracted forms (strong definites). This discovery then forms the starting point of a formal linguistic analysis.

In sum, the MDS methodology does not commit the researcher to one particular theoretical framework, and has indeed been used with a variety of theoretical frameworks. This includes classical typology as well as formal linguistics, indicating that these two traditions need not be incompatible or conflicting, but are in fact rather closely related when it comes to the study of cross-linguistic variation. \\

Some authors (e.g.\ \cite[18]{Georgakopoulos2018}) have claimed that whereas classical semantic maps are an \textit{explanans} (they constitute an explanation as they are the result of preceding cross-linguistic analysis), MDS maps can be seen as an \textit{explanandum}, i.e.\ they are visualizations of data that are not the end product, but the starting point of further linguistic analysis. This suggests a dichotomy that does not reflect the great diversity seen above of the types of applications that employ MDS maps for linguistic analysis. This leads to a more nuanced view in which MDS maps take up different intermediate positions in the explanatory process. Some MDS maps are indeed the starting point of further analysis, in particular in formal linguistic applications, as we detail in \S\ref{sec:MDSformal} below. In other settings, such as in language classification or lexical semantics research, MDS maps represent a classification of languages or forms. In that case, the maps themselves -- with interpretation of clusters and dimensions -- form the main object of analytic interest.

\subsubsection{MDS and formal paradigms}  \label{sec:MDSformal}
 
We want to zoom in a bit more on the situation in which parallel corpus data and an MDS analysis are used to build a formal analysis of a linguistic phenomenon. The general structure of a formal linguistic analysis starts with a body of empirical data, followed by the building of a model in a formal language (e.g.\ a logical or mathematical system of syntax or semantics) that can explain the observed data, and make novel predictions. Parallel corpus data coupled with an MDS analysis take the place of providing the empirical data that form the basis for the analysis. The advantage of the methodology is that it allows the researcher to recognize patterns in a large set of corpus data, which cannot be found by hand. As a result, the subsequent analysis will have a more comprehensive empirical coverage.

Looking at it this way, the different approaches to applying MDS can be appreciated by specifying the position that MDS maps take within the analytic process or process of argumentation. The classic typological papers use MDS maps to visualize cross-linguistic variation itself, and the dimensional/clustering patterns in the maps are the main theoretical interest, as this provides information about language classification. By contrast, the more formally oriented approaches have MDS maps in an earlier position within the analytic process: they use MDS to identify empirical distinctions that are relevant for building an analysis of the phenomenon in question. The MDS stage is then followed up by a formal analysis that proceeds in a manner that is fairly typical for the approach of formal linguistics. 

One potential confusion that may arise relates to the distinction between the theoretical basis for creating semantic maps and the theoretical paradigm for subsequent formal analysis. Several MDS papers are explicit about their assumptions regarding the theoretical basis of semantic map methodology. Starting in \textcite[\S2]{Walchli2010} and \textcite[\S3]{Walchli2012}, and later adopted in other MDS studies (e.g.\ \cite[167]{Swart2012}), a combination of \textit{exemplar semantics} and \textit{similarity semantics} has been proposed. This means that exemplars (individual occurrences) are compared instead of abstract concepts, and that similarity is a more basic notion than identity. The two are linked by Haiman's Isomorphism Hypothesis (``recurrent identity of form will always reflect some perceived similarity in communicative function''; \cite{Haiman1985}). This theoretical basis underlies MDS maps in which points represent individual contexts (see \S\ref{sec:contexts}). 


The theoretical debate about similarity as a foundation for building semantic maps should not be confused with theoretical assumptions that may be made relating to a formal analysis that is constructed based on data from MDS maps. Although MDS methodology and the resulting maps crucially rely on a notion of similarity between linguistic objects, it does not follow that conclusions drawn about the semantic content of these objects must be based on similarity rather than identity.

A case in point is \textcite{Klis2021}, who argue that variation in the domain of the \textsc{perfect} is to be described in terms of dynamic semantics, compositional semantics, lexical semantics, and other constraints. So, for them, using a similarity-based statistical technique to create maps does not prevent them from an analysis in terms of well-established paradigms from the tradition of formal linguistics.\\

In conclusion, this section addressed the theory-neutrality of the MDS method, by raising the question of whether MDS, in addition to a means to reveal descriptive patterns in complex multidimensional datasets, can be a valuable tool for theoretical linguists working in various paradigms. We have argued that this is the case: cross-linguistic comparison can be the starting point to, and the empirical core of, theoretical linguistic studies. Hence, multidimensional scaling on data from parallel corpora should be part of the linguist's toolkit. Within the area of formal linguistics, discussed in this subsection, the use of parallel corpus data is still a fairly recent advance in need of further development. In the next section, we discuss some potential directions of future work, which we hope will further integrate the use of parallel corpus data in developing (formal) linguistic theory, as well as point to alternatives for analysis through MDS.

\section{Future directions}\label{sec:future}

In this section, we point at two possible future directions for applying MDS in linguistic research. First, we describe how we can use MDS when compositionality comes into play (\S\ref{sec:composition}). So far, we have seen applications of MDS that only compare single lexical or grammatical features, but in most semantic domains, we see an interplay of variables. A compositional approach is therefore necessitated. 

Second, we cover some alternatives to MDS as a dimensionality reduction method (\S\ref{sec:alternatives}). Recently, techniques have surfaced that assign more weight to local rather than global variation. We show how these methods can yield different perspectives on the datasets at hand. 

\subsection{Lexical-compositional step} \label{sec:composition}

In most of the MDS work reviewed in this paper, the methodology has been applied to word-size or phrase-size units (motion verbs, tense forms, causatives, etc.), and comparison has been made based on one parameter. As a next step in the application of MDS techniques in semantic variation research, we envisage the application of this method to larger constructions (multiple words, or sentence-size), for which comparison would be made based on multiple parameters.

In abstract terms, consider a complex construction A whose meaning is compositionally determined by component expressions B and C:
\begin{center}
    \Tree [.A B C ]
\end{center}
One can study variation for B and C separately, and then make predictions for what variation for A looks like. Alternatively, one can take construction A as primary data, and annotate various grammatical properties of A, including properties that relate to B and C. Then, an MDS solution can be computed that considers these various parameters. This can be done either by a distance function that is a weighted average of distance measures for the various parameters (as described in \cite{Tellings-ZRH}), or by \textit{multi-mode} or \textit{multi-way} MDS, extensions of MDS that consider multiple similarity measures for each pair of objects \parencite{Leeuw2009}. We are not aware of the use of these MDS extensions in the linguistic domain, but they promise to provide a way to take advantage of the MDS methodology for studying variation in the meaning of complex constructions. In addition, they would allow for studying variation in meaning composition, which is one of the aims of semantic cross-linguistic research \parencite{Fintel2008}.

Two ongoing research projects in which this approach has been taken are \textcite{Swart2021} and \textcite{Tellings-ZRH}, both based on Europarl data. \Textcite{Swart2021} use MDS to study cross-linguistic variation in the compositional interaction between negation and the lexical choice of connective in NPI constructions such as English \textit{not\ldots until}. \textcite{Tellings-ZRH} investigates variation in the interaction of tense use and modal interpretations of conditional sentences.

\subsection{Alternatives to MDS}\label{sec:alternatives}

Dimensionality reduction methods are usually subdivided into those that attempt to retain global structure of the data, like MDS, and those that instead try to retain local structure, like Local Linear Embedding (LLE; \cite{Roweis2000}). Lastly, some methods aim to operate at both the global and the local level, e.g., t-Distributed Stochastic Neighbor Embedding (t-SNE; \cite{vanderMaaten2008}). In this section, we briefly compare these three kinds of algorithms and show their applications in linguistics.

Generally, the difference between global-first (or \textit{full spectral}) and local-first (\textit{sparse spectral}) methods is demonstrated using an artificial dataset called the Swiss roll, pictured in Figure \ref{fig:swissroll} below. In a true Swiss roll, a sponge cake is rolled up to create a distinctive swirl effect. Similarly, the data in this set are curved when taking a three-dimensional perspective, but flat from a two-dimensional perspective. 

\begin{figure}[!ht]
\begin{subfigure}{.5\textwidth}
\centering
\includegraphics[width=\linewidth]{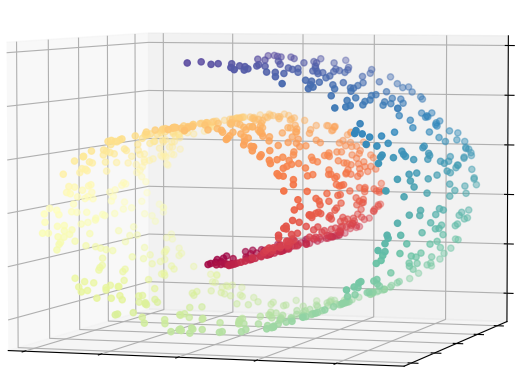}
\caption{Swiss roll dataset} \label{fig:swissroll}
\end{subfigure}
\begin{subfigure}{.5\textwidth}
\centering
\includegraphics[width=\linewidth]{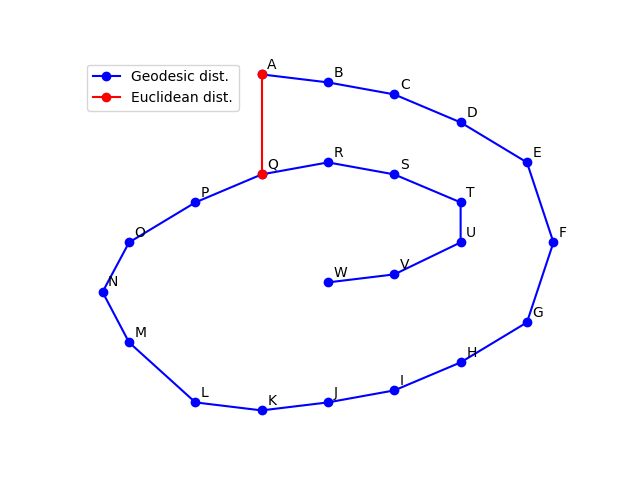}
\caption{Comparing distance functions}\label{fig:geodesic}
\end{subfigure}
\caption{Comparing Euclidean and geodesic distance through an artificial Swiss roll dataset.}
\end{figure}

Figure \ref{fig:geodesic} shows two ways of measuring distance between points in a two-dimensional rendering of the Swiss roll in Figure \ref{fig:swissroll}. On a global level, points $A$ and $Q$ are regarded close together, as their Euclidean distance is low. However, in the original sponge cake, $A$ and $Q$ would be rather far away, and only end up close after rolling up. So, when reducing the three-dimensional manifold to two dimensions, one preferably arrives at a solution that retains the adjacency of the points $A$, $B$, $C$, etc., rather than having $A$ end up close to $Q$. Local-first algorithms therefore attach more weight to geodesic distance instead: as a distance measure, they use the shortest path in terms of nearby points. So, the shortest path from $A$ to $Q$ goes via $B$, $C$, $D$, and so on. Hence, $A$ and $B$ end up close in the solution, while $A$ and $Q$ are far apart.

For the dataset pictured in Figure \ref{fig:swissroll}, MDS maps distant data points in the three-dimensional manifold to nearby points in the Cartesian plane. Consequently, as shown in Figure \ref{fig:comparison} below, MDS produces a rather similar two-dimensional output to the three-dimensional input data. Consequently, MDS fails to identify the underlying two-dimensional structure of the Swiss roll manifold.

LLE rather intends to retain local structure. As a result, LLE produces a low-dimensional solution that preserves the neighborhood of the manifold. For the Swiss roll dataset, the LLE output therefore resembles the two-dimensional structure of the manifold, as shown in Figure \ref{fig:comparison} below. A drawback to LLE is that the method has a general tendency to crowd points at the center of the map, which prevents gaps from forming between potential clusters \parencite[6-7]{vanderMaaten2008}. 

In Figure \ref{fig:comparison}, we find that t-SNE also keeps most of the local structure of the data intact, but we see some of the curvature of the original data as well. However, t-SNE additionally seems to have erroneously extracted two clusters, leading to displaying the dark blue points separately from the green points. Notably, t-SNE has parameters that require manual tuning, potentially hampering interpretation \parencite{Wattenberg2016}. Recently, Uniform Manifold Approximation and Projection (UMAP, \cite{McInnes2018}) entered the scene, claiming to preserve more global structure than t-SNE.

\begin{figure}[!ht]
\centering
\includegraphics[width=.9\textwidth]{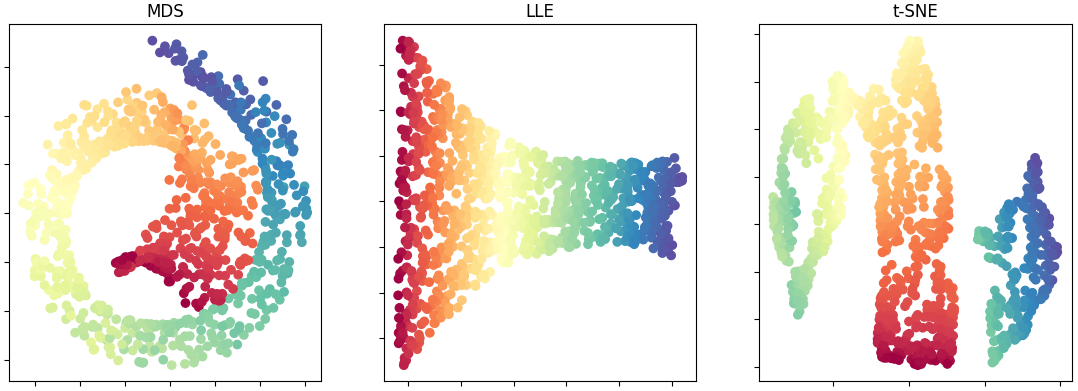}
\caption{Comparison of performance of three dimensionality reduction methods (MDS, LLE, and t-SNE) on the artificial Swiss roll dataset displayed in Figure \ref{fig:swissroll}.} 
\label{fig:comparison}
\end{figure}

Figure \ref{fig:comparison} seems to suggest that retaining local structure, like with LLE and t-SNE, yields better solutions regardless. However, whether the Swiss roll problem is at stake very much depends on the dataset at hand. Moreover, crucially, MDS can capture the main sources of cross-linguistic variation by assigning a linguistic interpretation to the resulting dimensions (see section \ref{sec:dimension}), while LLE and t-SNE are more suitable to identify clusters (see section \ref{sec:cluster}). Finally, importantly, all methods are susceptible to issues like the occurrence of horseshoe patterns as exposed in section \ref{sec:dimension} \parencite{Diaconis2008}. Effective dimensionality reduction hence requires an understanding of potential misinterpretations \parencite{Nguyen2019}.

While MDS prevails as the main method used in (typological) linguistics, recent research has shown applications of t-SNE and UMAP. For example, \textcite{Asgari2017} apply t-SNE to cross-linguistic variation in tense markers. They show how grammatical markers, e.g.\ past tense marking with \textit{ti} in Seychellois Creole, can function as pivots to find all past-referring contexts in the parallel corpus of Bible translations. Iteratively selecting more pivots, e.g.\ Fijian \textit{qai}, then allows to discern sub-types of past-referring contexts: \textit{qai} is used as a past tense marker in narrative progression, but not in progressive or modal contexts. Applying t-SNE on formal similarity in the parallel corpus as a whole then neatly shows clustering of these aforementioned functions in the domain of past reference. Another example is \textcite{Georgakopoulos2022}, who apply UMAP in the semantic domains of perception and cognition by comparing colexification patterns across languages. They find that cross-linguistically, verbs almost never colexify \textsc{hear}, \textsc{see}, \textsc{think (believe)}, and \textsc{learn} with each other, as these meanings are found in completely separate areas on the map generated through UMAP. On the other hand, the meanings \textsc{understand} and \textsc{know (something)} are frequently colexified, and as a result, verbs expressing these meanings are mostly found in the same cluster on the resulting map.

While we are unaware of implementations of LLE to (re)generate semantic maps, this section, along with sections \ref{sec:dimension} and \ref{sec:cluster}, shows that multiple visualizations are often required to arrive at a full interpretation (cf.\ \cite[50]{Cysouw2008}, \cite{Georgakopoulos2022}): we generally do not know the underlying structure of our dataset.

\section{Conclusion}\label{sec:conclusion}

This paper reviewed how multidimensional scaling is used to create semantic maps in linguistic typology and cross-linguistic semantics. We have seen that MDS stands for a collection of algorithms that can reduce the dimensionality of a highly complex dataset, and represent this visually. Starting with a notion of similarity between linguistic objects, applying MDS results in a visualization of both cross-linguistic variation and single-language patterns, which then can be used to answer a variety of linguistic research questions.

What makes reading the MDS literature in linguistics potentially difficult is that there is so much variation with respect to various parameters of MDS implementations. These parameters include the particular MDS algorithm that is used, the type of linguistic data used as input, the similarity measure between primitives, what the points on the map represent, how clusters and dimensions are interpreted, and the place that MDS maps occupy in the process of linguistic argumentation. By identifying and explaining these parameters in this paper, and introducing useful terminology for describing MDS studies (\textit{map coloring}, \textit{dimension interpretation}, \textit{cluster interpretation}, etc.), we hope to have provided the means to make existing MDS-based work in linguistics more accessible.

At the same time, we hope this paper will prompt future MDS studies. We suggested two directions for future work in particular. First, the use of MDS in a setting in which multiple semantic features are at play in a compositional way, so that the MDS methodology can contribute to the study of cross-linguistic variation of compositional structures. Second, we discussed how MDS can be complemented and compared with other dimensionality reduction techniques.

\begin{acknowledgement}
  We thank Henriëtte de Swart, Bert Le Bruyn, Martín Fuchs, Chou Mo, Jianan Liu, and Joost Zwarts for their valuable feedback on earlier versions of this paper. We thank three anonymous reviewers for their very detailed and helpful comments that helped us improve the paper. All remaining errors are ours.
  
  We are grateful to Bernhard W\"alchli for sharing data from the \textcite{Walchli2012} paper that allowed us to create Figure \ref{map:WC}, and to Natalia Levshina for sharing data from her \citeyear{Levshina2020} paper that allowed us to create Figure \ref{map:hca}. We thank the editors of \textit{Studies in Language} for granting permission to reproduce Figure \ref{map:hartmann} from the original in \textcite[Figure 5]{Hartmann2014}. We thank the editors of \textit{Baltic Linguistics} for granting permission to reproduce Figure \ref{map:pam} from the original in \textcite[Figure 2]{Walchli2018}.
\end{acknowledgement}

\begin{funding}
  This publication is part of the project \textit{Time in Translation} (with project number 360-80-070) of the research programme Free Competition Humanities which is financed by the Dutch Research Council (NWO).
\end{funding}

\printbibliography

\end{document}